\documentclass[letterpaper]{article} 
\usepackage{aaai25}  
\usepackage{times}  
\usepackage{helvet}  
\usepackage{courier}  
\usepackage[hyphens]{url}  
\usepackage{graphicx} 
\usepackage{natbib}  
\usepackage{caption} 
\usepackage{algorithm}
\usepackage{algorithmic}
\usepackage[]{multirow}
\usepackage{color}
\usepackage{xcolor} 
\usepackage{colortbl}
\usepackage{amssymb}
\usepackage{subfigure}
\usepackage{amsmath}
\usepackage{xspace}
\definecolor{cvprblue}{rgb}{0.21,0.49,0.74}

\makeatletter
\DeclareRobustCommand\onedot{\futurelet\@let@token\@onedot}
\def\@onedot{\ifx\@let@token.\else.\null\fi\xspace}
\def\eg{\emph{e.g}\onedot} 
\def\ie{\emph{i.e}\onedot}

\makeatother

\usepackage{newfloat}
\usepackage{listings}
\DeclareCaptionStyle{ruled}{labelfont=normalfont,labelsep=colon,strut=off} 
\floatstyle{ruled}
\newfloat{listing}{tb}{lst}{}
\floatname{listing}{Listing}
\title{ZeroMamba: Exploring Visual State Space Model for Zero-Shot Learning}
\author{
    Wenjin Hou\textsuperscript{\rm 1},
    Dingjie Fu\textsuperscript{\rm 2},
    Kun Li\textsuperscript{\rm 1},
    Shiming Chen\textsuperscript{\rm 2,3},
    Hehe Fan\textsuperscript{\rm 1}\thanks{Corresponding author},
    Yi Yang\textsuperscript{\rm 1}
}
\affiliations{
    \textsuperscript{\rm 1}ReLER Lab, Zhejiang University, China\\
    \textsuperscript{\rm 2}Huazhong University of Science and Technology (HUST), China \qquad
    \textsuperscript{\rm 3}Mohamed bin Zayed University of AI\\
    
    {\{wenjinhou, hehefan, yangyics\}}@zju.edu.cn
}

\usepackage{bibentry}

\begin{document}

\maketitle

\begin{abstract}
Zero-shot learning (ZSL) aims to recognize unseen classes by transferring semantic knowledge from seen classes to unseen ones, guided by semantic information. To this end, existing works have demonstrated remarkable performance by utilizing global visual features from Convolutional Neural Networks (CNNs) or Vision Transformers (ViTs) for visual-semantic interactions. Due to the limited receptive fields of CNNs and the quadratic complexity of ViTs, however, these visual backbones achieve suboptimal visual-semantic interactions. In this paper, motivated by the visual state space model (\ie, Vision Mamba), which is capable of capturing long-range dependencies and modeling complex visual dynamics, we propose a parameter-efficient ZSL framework called \textbf{ZeroMamba} to advance ZSL. Our ZeroMamba comprises three key components: Semantic-aware Local Projection (SLP), Global Representation Learning (GRL), and Semantic Fusion (SeF). Specifically, SLP integrates semantic embeddings to map visual features to local semantic-related representations, while GRL encourages the model to learn global semantic representations. SeF combines these two semantic representations to enhance the discriminability of semantic features. We incorporate these designs into Vision Mamba, forming an end-to-end ZSL framework. As a result, the learned semantic representations are better suited for classification. Through extensive experiments on four prominent ZSL benchmarks, ZeroMamba demonstrates superior performance, significantly outperforming the state-of-the-art (\ie, CNN-based and ViT-based) methods under both conventional ZSL (CZSL) and generalized ZSL (GZSL) settings. Code is available at: \textit{\textcolor{red}{https://anonymous.4open.science/r/ZeroMamba}}.
\end{abstract}

\section{Introduction}
\begin{figure}[ht]
  \centering
   \includegraphics[width=1.0\linewidth]{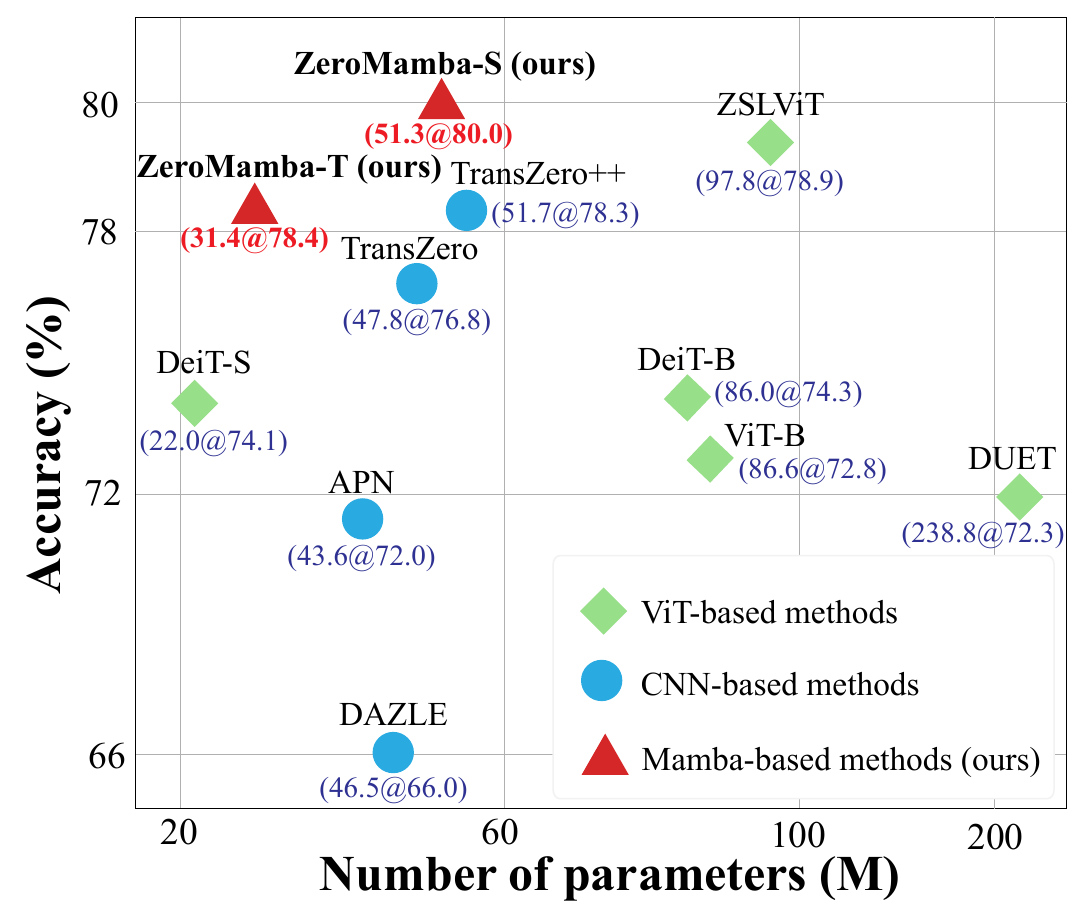}
   \caption{Compared with state-of-the-art CNN-based and ViT-based methods on CUB, our proposed ZeroMamba achieves the best trade-off between \textbf{Accuracy} and \textbf{Number of parameters}.}
   \label{fig:motivate}
\end{figure}

Zero-shot learning (ZSL) recognizes unseen classes with the support of shared semantic information (\eg, category attributes \cite{lampert2009learning, lampert2013attribute}, semantic embeddings \cite{huynh2020fine, radford2021learning}). Since images of unseen classes are unavailable during training, mainstream methods \cite{huynh2020fine,wang2021dual,xu2022attribute,chen2022msdn,chen2023evolving,chen2024progressive,chen2024ral, Hou_2024_CVPR,naeem2024i2dformer+} focus on establish visual-semantic interactions to align visual and semantic features, transferring semantic knowledge from seen classes to unseen ones. As a result, learning discriminative visual and semantic representations is crucial for ZSL.

Current methods typically leverage visual backbones with the powerful learning capabilities of CNNs (\eg, ResNet-101 \cite{he2016deep}) or ViTs (\eg, ViT \cite{dosovitskiy2020image}) to extract visual features. However, CNNs suffer from limited receptive fields in convolution operations \cite{luo2016understanding}, and ViTs face quadratic complexity issues in self-attention calculations \cite{habib2023knowledge}. Additionally, these visual backbones fail to learn desirable visual-semantic correspondences due to the lack of semantic guidance during pre-training on ImageNet \cite{russakovsky2015imagenet}. Given these limitations, the performance and computational efficiency of CNN-based and ViT-based ZSL methods remain suboptimal, as illustrated in Fig. \ref{fig:motivate}.

Recently, the visual state space model (\ie, Vision Mamba) \cite{zhu2024vision,huang2024localmamba}, a novel and promising network architecture with advantages in linear complexity and global receptive fields, has demonstrated remarkable performance across various visual perception tasks, such as image \cite{liu2024vmamba}, video \cite{li2024videomamba,park2024videomamba} and multi-modal \cite{fu2024mambagesture} scenarios. Drawing inspiration from its ability to capture long-range dependencies and model complex visual dynamics, we \textit{explore the visual state space model} to advance ZSL. To this end, the basic idea is straightforward: utilize Vision Mamba as the backbone to extract visual features and then directly map these features to the semantic space via a multi-layer perceptron (MLP) network for nearest-neighbor matching. However, significant challenges for ZSL lie in effectively bridging visual-semantic spaces and inserting discriminative semantic information. The basic solution tends to overfit to seen classes, severely limiting the transfer of intrinsic semantic knowledge. Thus, improving Visual Mamba to better suit ZSL and provide valuable insights are crucial issues.

In light of the above observations, our major innovations in this paper are \textbf{three} effective designs: \textit{Firstly}, we introduce a Semantic-aware Local Projection (SLP) module conditioned on semantic embeddings to learn semantic-related local representations; \textit{Secondly}, we develop a Global Representation Learning (GRL) module to bridge the visual and semantic spaces; \textit{Thirdly}, we devise a Semantic Fusion (SeF) strategy to combine the two semantic representations, enhancing the discriminability of semantic features. By incorporating these three simple yet effective designs into Vision Mamba, we form an end-to-end ZSL framework called \textbf{ZeroMamba}. This unified framework allows us to simultaneously optimize visual and semantic representations during training, significantly boosting effective visual-semantic interactions. As shown in Fig. \ref{fig:motivate}, our ZeroMamba is a parameter-efficient method that achieves the best trade-off between accuracy and the number of parameters. Overall, our contribution can be summarized as follows:

\begin{itemize}
\item We propose ZeroMamba, a parameter-efficient Mamba-based framework for ZSL that exhibits exceptional generalization of unseen classes and meanwhile has the advantages of global receptive fields. To the best of our knowledge, ZeroMamba is the first work attempting to explore Vision Mamba for the ZSL task.
\item To adapt to ZSL, we design the Semantic-aware Local Projection (SLP) module, the Global Representation Learning (GRL) module, and the Semantic Fusion (SeF) strategy,  integrating them into Vision Mamba to form an end-to-end ZSL framework.
\item Extensive experiments with analyses on three prominent ZSL benchmarks and one large-scale dataset show that ZeroMamba sets a new state-of-the-art (SOTA) in both CZSL and GZSL settings, illustrating the effectiveness and superiority of our approach.
\end{itemize}

\section{Related Work}
\noindent{\textbf{Zero-Shot Learning.}} 
Zero-shot learning (ZSL) transfers semantic knowledge from seen classes to unseen ones through visual-semantic interactions. According to the direction of visual and semantic mappings, mainstream ZSL methods can be categorized into generative and embedding-based approaches \cite{xian2019f}, implemented via semantic$\rightarrow$visual and visual$\rightarrow$semantic mappings, respectively. Typically, visual features are extracted using visual backbones like CNNs or ViTs. Semantic information is derived from category attributes. In this regard, enhancing the representation capability of these features has garnered widespread attention. In CNN feature-based methods, early works \cite{huynh2020fine,chen2021hsva,chen2021free,han2021contrastive,cetin2022closed,feng2022non} directly utilized global visual features, while recent studies \cite{wang2021dual,xu2022attribute,chen2022transzero,chen2024ral, li2024improving} focus on local discriminative features. In these works, APN \cite{xu2020attribute} and TransZero \cite{chen2022transzero} obtain semantic-related representations through local attention mechanisms; DPPN \cite{wang2021dual} simultaneously updates attributes and class prototypes; CE-GZSL \cite{han2021contrastive} and ICCE \cite{kong2022compactness} map visual features to latent spaces; DSECN \cite{li2024improving} explores diverse semantics from external class names. 

Recently, ViTs with excellent self-attention mechanisms have also been applied to ZSL. For example, DUET \cite{chen2023duet} introduces cross-modal masks; PSVMA \cite{liu2023progressive} proposes a semantic-visual mutual adaption network; ZSLViT \cite{chen2024progressive} leverages semantic information to guide network learning; CVsC \cite{chen2024causal} learns substantive visual-semantic correlations. In addition, large-scale vision-language models (\eg, CLIP \cite{radford2021learning}) emerge with impressive ZSL transfer capabilities by aligning visual-semantic features in a common space. Analyzing these methods reveals that discriminative visual features and semantic representations are crucial for effective visual-semantic interactions.\\

\noindent{\textbf{Vision Mamba.}}
State space models (SSMs), particularly Mamba \cite{mamba,mamba2}, have recently demonstrated strong long-range modeling capabilities while maintaining linear complexity. In the field of Computer Vision (CV), Vim \cite{zhu2024vision} is the first Mamba-based backbone network, showcasing superior performance and the ability to capture complex visual dynamics across a variety of vision tasks (\eg, image classification, semantic segmentation and object detection). Subsequent Mamba-based models, such as VMamba \cite{liu2024vmamba} and LocalMamba \cite{huang2024localmamba}, primarily focus on image scanning strategies and Mamba blocks. VideoMamba \cite{li2024videomamba,park2024videomamba} extends Mamba to the video domain. However, Mamba-based ZSL methods have yet to be explored. Therefore, this work presents a simple Mamba-based framework specifically for ZSL and provides a comprehensive analysis of its performance. 

\section{Proposed Method}
\begin{figure*}[ht]
  \centering
   \includegraphics[width=1.0\linewidth]{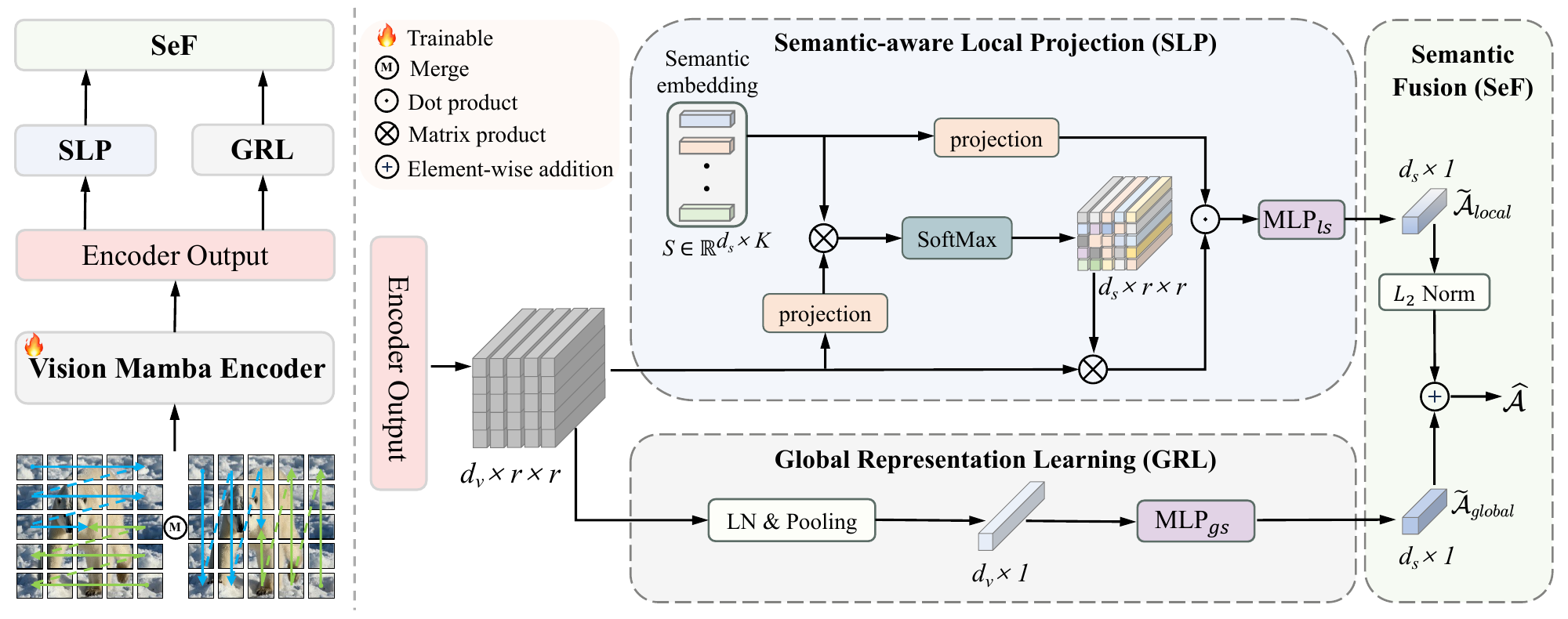}
   \caption{\textbf{Left:} The framework of \textbf{ZeroMamba}. Image patches are traversed along four-way scanning and fed into the Vision Mamba Encoder. The major innovation of our work lies in three simple yet effective designs: the SLP module, the GRL module, and the SeF strategy. By incorporating these designs into vanilla VMamba \cite{liu2024vmamba}, we form an end-to-end framework capable of robust ZSL. \textbf{Right:} The structural details of the SLP module, the GRL module, and the SeF strategy.}
   \label{fig:framework}
\end{figure*}
ZeroMamba aims to incorporate the visual state space model (\ie, Vision Mamba) into ZSL. Fig. \ref{fig:framework} illustrates the framework of our ZeroMamba. In this section, we first present the definition of the ZSL task. Next, we introduce the preliminaries of the state space models and the selection mechanism. Then, we describe the details of the proposed method. Finally, we demonstrate the model optimization and zero-shot prediction process.\\
\noindent{\textbf{Problem Definition.}} Assume $\mathcal{D}^s=\{(x_i,y_i)|x_i\in{\mathcal{X}^s},y_i\in{\mathcal{Y}^s}\}$ with $\mathcal{C}^s$ classes, where $x_i$ and $y_i$ represent the image of the \textit{i}-th sample and its label, respectively. A disjoint set $\mathcal{D}^u=\{(x_i,y_i)|x_i\in{\mathcal{X}^u},~y_i\in{\mathcal{Y}^u}\}$ consists of $\mathcal{C}^u$ classes. Notably, $\mathcal{C}^s\cap \mathcal{C}^u=\emptyset$. In the conventional zero-shot learning (CZSL) setting, the training set $\mathcal{D}^{tr}$ is a subset of $\mathcal{D}^s$. The main goal is to classify a test image $x\in{\mathcal{D}^u}$ that belongs to an unseen class, given that there are no training images available for unseen classes. In the generalized zero-shot learning (GZSL) setting, the testing samples include both seen and unseen classes, \ie, ${c\in\mathcal{C}^s\cup\mathcal{C}^u=\mathcal{C}}$. In this paper, we utilize class-level attribute vectors (\ie, semantic prototypes) $\mathcal{A}\in\mathbb{R}^{K\times{C}}$ to bridge the gap between seen and unseen classes, where $K$ refers to the number of attributes and $C$ denotes the total number of categories. We also use the shared semantic embeddings $\mathcal{S}\in\mathbb{R}^{d_s\times{K}}$, extracted using GloVe, to boost visual-semantic interactions.

\subsection{Preliminaries}
\noindent{\textbf{State Space Models}}. 
State space models (SSMs) have emerged as a novel alternative to CNNs and ViTs for modeling long-range dependency. Continuous-time SSMs map an input stimulation $x(t)\in\mathbb{R}$ to a response $y(t)\in\mathbb{R}$ via a hidden state $h(t)\in\mathbb{R}^N$. This progress can be mathematically formulated as follows:
\begin{equation}
\label{eq.1}
\begin{aligned}
h'(t)&=\textbf{\textit{A}}h(t)+\textbf{\textit{B}}x(t),\\ y(t)&=\textbf{\textit{C}}h(t),
\end{aligned}
\end{equation}
where $\textbf{\textit{A}}\in\mathbb{R}^{N\times{N}}$, $\textbf{\textit{B}}\in\mathbb{R}^{N\times{1}}$ and $\textbf{\textit{C}}\in\mathbb{R}^{1\times{N}}$ are the projection parameters. The term $h'(t)$ denotes the derivative of $h(t)$ with respect to time $t$.

To handle discrete input sequences, the commonly used zero-order hold (ZOH) assumption leverages a time step parameter $\Delta$ to convert the continuous parameters $\textbf{\textit{A}}$ and $\textbf{\textit{B}}$ into their discrete counterparts $\overline{\textbf{\textit{A}}}$ and $\overline{\textbf{\textit{B}}}$ as follows:
\begin{equation}
\label{eq.2}
\begin{aligned}
&\overline{\textbf{\textit{A}}}=\text{exp}(\Delta\textbf{\textit{A}}),\\ &\overline{\textbf{\textit{B}}}=(\Delta\textbf{\textit{A}})^{-1}(\text{exp}(\Delta\textbf{\textit{A}})-\textbf{\textit{I}})\cdot{\Delta\textbf{B}}.
\end{aligned}
\end{equation}
After discretizing, the formulation of Eq. (\ref{eq.1}) is altered to:
\begin{equation}
\label{eq.3}
\begin{aligned}
h_t&=\overline{\textbf{\textit{A}}}h_{t-1}+\overline{\textbf{\textit{B}}}x_t,\\ y_t&=\textbf{\textit{C}}h_t.
\end{aligned}
\end{equation}
The model utilizes Eq. (\ref{eq.3}) to facilitate efficient inference, where the inputs are processed sequentially. For efficient and parallelizable training, Eq. (\ref{eq.3}) can be reformulated and computed as a global convolution:
\begin{equation}
\label{eq.4}
\begin{aligned}
\overline{\textbf{\textit{K}}}&=(\textbf{\textit{C}}\overline{\textbf{\textit{B}}},\textbf{\textit{C}}\overline{\textbf{\textit{AB}}},...,\textbf{\textit{C}}\overline{\textbf{\textit{A}}}^{L-1}\overline{\textbf{\textit{B}}}), \\ y&=x\ast\overline{\textbf{\textit{K}}},
\end{aligned}
\end{equation}
where $L$ represents the length of the input sequence, and $\overline{\textbf{\textit{K}}}\in\mathbb{R}^L$ is the convolutional kernel.\\

\noindent{\textbf{Selection Mechanism}}. 
Recent SSMs, such as Mamba models \cite{mamba,mamba2}, leverage an input-dependent selection mechanism to address the limitations of fixed parameterization. By adopting such a mechanism, Mamba models exhibit linear scalability and demonstrate strong capabilities in long-range modeling.

In vision tasks, the Mamba model designed for 1-D sequence may not be optimal for handling 2-D image inputs. Therefore, Vision Mamba models have introduced a variety of selective scan mechanisms. Concretely, Vim \cite{zhu2024vision} combines Mamba with bidirectional SSM paths, VMamba \cite{liu2024vmamba} proposes the 2D-Selective-Scan method (SS2D), LocalMamba \cite{huang2024localmamba} segments the input image into local windows to perform SSM in different directions while preserving global SSM operations. In our work, we use SS2D to merge contextual information from four directions, following the design of VMamba.\\

\noindent{\textbf{Overview.}}
Following the embedding-based ZSL methods, we can achieve ZSL through the following steps: First, use a pre-trained Vision Mamba model as the visual backbone to extract visual feature $\boldsymbol{v}_i$ of the image $x_i$; Then, apply a multi-layer perceptron (MLP) with a parameter matrix $\boldsymbol{W}\in\mathbb{R}^{d_s\times{d_v}}$ as a semantic predictor to obtain a global semantic feature $\boldsymbol{a}_i$ for classification:
\begin{equation}
\label{eq.5}
\boldsymbol{a}_i = f(\boldsymbol{v}_i) = \boldsymbol{W}\boldsymbol{v}_i.
\end{equation}

However, such a naive method inevitably leads to overfitting to seen classes, severely limiting the transfer of intrinsic semantic knowledge.

To tackle the above challenges, we proposed a Mamba-based ZSL framework, named ZeroMamba,  which is shown in Fig. \ref{fig:framework}.
ZeroMamba first acquires image patches through a four-way selective scan mechanism to aggregate contextual information. These patches are then fed into the Vision Mamba Encoder to extract visual features (due to page limits, the details of the Vision Mamba Encoder are provided in the Appendix \textcolor{red}{A}. More details can also be found in VMamba \cite{liu2024vmamba}.). Next, three components -- SLP module, GRL module, and Semantic Fusion strategy -- are combined with the Vision Mamba Encoder. As a result, ZeroMamba forms an end-to-end framework that integrates semantic information into the network and simultaneously optimizes visual and semantic representations. The following sections detail our proposed designs.

\subsection{Semantic-aware Local Projection (SLP)}
To enhance ZeroMamba with rich semantic information, we propose a Semantic-aware Local Projection (SLP) module. As shown in Fig. \ref{fig:framework}, the SLP first extracts semantic embeddings: $\mathcal{S}=\{\boldsymbol{s}_i\}_{i=1}^{K}$, where $\boldsymbol{s}_i\in\mathbb{R}^{d_s}$ denotes the semantic embedding of the $i$-th attribute. Subsequently, we flatten the encoder's output feature map (\ie, $\mathcal{F}\in\mathbb{R}^{d_v\times{r}\times{r}}$$\rightarrow$$\boldsymbol{F}_{r'}\in\mathbb{R}^{d_v\times{r'}}$), where $r'$ corresponds to $r\times{r}$ image regions. For these features, we apply two projection parameters $\boldsymbol{W}_1$, $\boldsymbol{W}_2$ to calculate the attention map $\mathcal{M}$ and obtain the local semantic representation $f_s\in\mathbb{R}^{d_s\times{1}}$. This process is formalized as follows:
\begin{equation}
\label{Eq:SLP}
\mathcal{M} = \text{SoftMax}(\mathcal{S}{\boldsymbol{W}_1}{\boldsymbol{F}_{r'}}),~f_s = \mathcal{S}{\boldsymbol{W}_2}\odot({{\mathcal{M}}{\boldsymbol{F}_{r'}}})^T,
\end{equation}
where $\odot$ denotes dot product operation, and ${{\mathcal{M}}{\boldsymbol{F}_{r'}}}$ represents the semantic-related attention feature. 

Upon obtaining $f_s$, we employ an MLP (\ie, $\text{MLP}_{ls}$) that maps local semantic representations to semantic space, denoted as $\widetilde{\mathcal{A}}_{local}$:
\begin{equation}
\label{Eq:alocal}
\begin{aligned}
\widetilde{\mathcal{A}}_{local} = \text{MLP}_{ls}(f_s).
\end{aligned}
\end{equation}

Through this process, the SLP effectively captures local semantic-related representations, facilitating visual-semantic interactions.

\subsection{Global Representation Learning (GRL)}
We devise a Global Representation Learning (GRL) module to discover global semantic information and bridge visual and semantic spaces. As shown in Fig. \ref{fig:framework}, the GRL employs LayerNorm to stabilize feature activations and subsequently uses average pooling to obtain an informative global visual feature $f_v\in\mathbb{R}^{d_v\times{1}}$. At last, GRL leverages a MLP (\ie, $\text{MLP}_{gs}$) to project the visual embedding to global semantic representation, denoted as $\widetilde{\mathcal{A}}_{global}$:
\begin{equation}
\label{Eq:aglocal}
\begin{aligned}
\widetilde{\mathcal{A}}_{global} = \text{MLP}_{gs}(f_v).
\end{aligned}
\end{equation}

\subsection{Semantic Fusion Strategy (SeF)}
By applying SLP and GRL, we obtain global and local semantic representations. We argue that fusing these representations enhances the exploration of complementary semantic information for classification. To this end, we propose a simple Semantic Fusion (SeF) strategy, formulated as follows:
\begin{equation}
\label{eq.7}
\widehat{\mathcal{A}} = \widetilde{\mathcal{A}}_{global} + \text{Norm}(\widetilde{\mathcal{A}}_{local}),
\end{equation}
where Norm denotes the $L_{2}$ normalization, which suppresses the significance of $\widetilde{\mathcal{A}}_{local}$. 
To ensure alignment with the semantic prototypes $\mathcal{A}$, we introduce a semantic constraint loss $\mathcal{L}_{sc}$ to guide the optimization of the semantic representation $\widehat{\mathcal{A}}$, defined as:
\begin{equation}
\label{Eq:sc_loss}
\mathcal{L}_{sc} = \mathbb{E}[\|\widehat{\mathcal{A}}-\mathcal{A}\|_1].
\end{equation}

Based on the above fusion operations, we optimize both global and local semantic representations, boosting effective visual-semantic interactions.

\subsection{Model Optimization and Zero-Shot Prediction}
\noindent{\textbf{Optimization}}. In addition to the semantic constraint loss mentioned above, we optimize the cross-entropy loss between the prediction and ground-truth label:
\begin{equation}
\mathcal{L}_{ce} = -\log\frac{\text{exp}(p^c_i)}{\sum_{c'\in\mathcal{C}^s}{\text{exp}(p_i^{c'})}},
\end{equation}
where $p^c_i$ is the probability that image $i$ belongs to class $c$, formulated as:
\begin{equation}
p^c_i = cos(\widehat{\mathcal{A}^c_i},~\mathcal{A}^c).
\end{equation}
Here, $cos(\cdot)$ denotes the cosine function, which measures the similarity between predicted semantic feature $\widehat{\mathcal{A}}$ and class semantic prototypes $\mathcal{A}$.

Finally, the overall loss function for our model is:
\begin{equation}
\mathcal{L} = \mathcal{L}_{ce} + \lambda_{sc}\mathcal{L}_{sc},
\end{equation}
where $\lambda_{sc}$ is a hyper-parameter to control the magnitude of $\mathcal{L}_{sc}$, with its value varying across different datasets. \\

\noindent{\textbf{Prediction}}. After training, we input test sample $x_i$ into ZeroMamba, obtain its semantic representation $\widehat{\mathcal{A}_i}$, and calculate the probabilistic vector $p_i$. To this end, we predict the class label $y^{\ast}$ using the following formulation:
\begin{equation}
y^{\ast} = \mathop{\text{arg}\ \text{max}}\limits_{y\in\mathcal{Y}^{u} /\mathcal{Y}^s\cup\mathcal{Y}^u}
(p_i+\lambda_{col}\mathbb{I}_{[y\in\mathcal{Y}^u]}).
\end{equation}
$\mathbb{I}_{[y\in\mathcal{Y}^u]}$ is an indicator function, which is 1 when $y\in\mathcal{Y}^u$ and 0 otherwise. $\lambda_{col}$ is a calibration coefficient that explicitly calibrates the sensitivity of the predictor to unseen classes, and $\mathcal{Y}^{u} /\mathcal{Y}^s\cup\mathcal{Y}^u$ corresponds to the CZSL/GZSL setting.

\begin{table*}[htbp]
  \centering
  \resizebox{\linewidth}{!}{
  \begin{tabular}{l|c|c|c|ccc|c|ccc|c|ccc}
  \hline
      \multirow{3}{*}{\textbf{Method}}  &\multirow{3}{*}{\textbf{Backbone}}&\multirow{3}{*}{\textbf{Venue}}&\multicolumn{4}{c|}{\textbf{CUB}}&\multicolumn{4}{c|}{\textbf{SUN}}&\multicolumn{4}{c}{\textbf{AWA2}}\\
      \cline{4-15}
      &&&\multicolumn{1}{c|}{CZSL} & \multicolumn{3}{c|}{GZSL}&\multicolumn{1}{c|}{CZSL} & \multicolumn{3}{c|}{GZSL}&\multicolumn{1}{c|}{CZSL} & \multicolumn{3}{c}{GZSL}\\
      \cline{4-15} 
      &&&\multicolumn{1}{c|}{\rm{Acc}} & \rm{U} & \rm{S} &\rm{H} & \multicolumn{1}{c|}{\rm{Acc}} & \rm{U} &\rm{S} & \rm{H} & \multicolumn{1}{c|}{\rm{Acc}} &\rm{U}  & \rm{S}  & \rm{H}\\
      \hline
      \multicolumn{15}{c}{\textbf{CNN-based methods}}\\
      \hline
      DAZLE \cite{huynh2020fine} & ResNet-101 & CVPR'20 &66.0&56.7&59.6&58.1&59.4&52.3&24.3&33.2&67.9&60.3&75.7&67.1\\
      APN \cite{xu2020attribute} & ResNet-101 & NeurIPS'20 &72.0&65.3& 69.3&67.2&61.6& 41.9&34.0&37.6&68.4&57.1&72.4&63.9\\
      HSVA \cite{chen2021hsva} & ResNet-101 & NeurIPS'21 & 62.8 &52.7 &58.3& 55.3& 63.8& 48.6 &39.0& 43.3 &--& 59.3& 76.6 &66.8\\
      SE-GZSL \cite{kim2022semantic} & ResNet-101 & AAAI'22&--& 53.1 &60.3& 56.4& --& 45.8& 40.7& 43.1& --& 59.9& 80.7& 68.8\\
      TransZero++ \cite{chen2022transzero++} & ResNet-101 & TPAMI'22&78.3& 67.5& 73.6& 70.4 &67.6 &48.6 &37.8 &42.5 &72.6 &64.6 &82.7 &72.5\\
      FREE + ESZSL \cite{cetin2022closed} & ResNet-101 & ICLR'22 & --& 51.6& 60.4& 55.7&--&48.2& 36.5& 41.5&--&51.3& 78.0& 61.8\\
      f-CLSWGAN + DSP \cite{chen2023evolving} & ResNet-101& ICML'23 & --&51.4& 63.8& 56.9&--&48.3 &43.0 &45.5&--&60.0& 86.0& 70.7\\
      CDL + OSCO \shortcite{cavazza2023no} &ResNet-101& TPAMI'23&--&29.0& 69.0& 40.6&--&32.0& \textbf{\color{red}65.0}& 42.9&--&48.0& 71.0& 57.1\\
      ICIS \cite{christensen2023image}&ResNet-101&ICCV'23&60.0&45.8& 73.7& 56.5&51.8&45.2& 25.6& 32.7&64.6&35.6&\textbf{\color{blue} 93.3}& 51.6 \\
      HAS \cite{chen2023zero} &ResNet-101 & ACM MM'23 & 76.5& 69.6& 74.1& 71.8& 63.2& 42.8& 38.9& 40.8& 71.4& 63.1& 87.3& 73.3\\
      TransZero + ALR \cite{chen2024ral} & ResNet-101&SCIS'24&78.8&\textbf{\color{blue}70.4}& 69.0& 69.7&66.2&\textbf{\color{blue}52.7}& 34.0& 41.3&71.2& 61.6& 82.3& 70.5\\
      DSECN \cite{li2024improving} &ResNet-101&CVPR'24&40.9&--&--&45.3&49.1&--&--&38.5&40.0&--&--&53.7\\
      \hline
      \multicolumn{15}{c}{\textbf{ViT-based methods}}\\
      \hline
      ViT-ZSL \cite{alamri2021multi}& ViT-Large&IMVIP'21 &--& 67.3& 75.2& 71.0&--& 44.5& \textbf{\color{blue}55.3} &\textbf{\color{red}49.3}&-- &51.9& 90.0& 68.5\\
      CLIP \cite{radford2021learning} & ViT-Base & ICML'21 & --& 55.2 &54.8& 55.0&--&--&--&--&--&--&--&--\\
      DUET \cite{chen2023duet} & ViT-Base & AAAI'23 & 72.3 &62.9 &72.8 &67.5& 64.4& 45.7& 45.8& 45.8& 69.9& 63.7& 84.7& 72.7\\
      TFVAEGAN + SHIP \cite{wang2023improving}& ViT-Base &ICCV'23 & --&21.1& \textbf{\color{red}84.4}& 34.0&--&--&--&--&--&43.7 &\textbf{\color{red}96.3}& 60.1\\
      I2MVFormer-Wiki \cite{naeem2023i2mvformer} &ViT-Base& CVPR'23&42.1&32.4 &63.1 &42.8&--&--&--&--&\textbf{\color{blue}73.6}&66.6& 82.9& 73.8\\
      I2DFormer+ \cite{naeem2024i2dformer+} & ViT-Base & IJCV'24 & 45.9&38.3& 55.2& 45.3&--&--&--&--&\textbf{\color{red}77.3}&\textbf{\color{red}69.8}& 83.2& \textbf{\color{blue}75.9}\\
      ZSLViT \cite{chen2024progressive} & ViT-Base & CVPR'24 & \textbf{\color{blue}78.9}& 69.4& \textbf{\color{blue}78.2}& \textbf{\color{blue}73.6}&\textbf{\color{blue} 68.3}& 45.9& 48.4& 47.3& 70.7& 66.1& 84.6& 74.2\\
      \hline
      \multicolumn{15}{c}{\textbf{Mamba-based method}}\\
      \hline
      \rowcolor{gray!20}
       \textbf{ZeroMamba (Ours)}& VMamba-Small &--&\textbf{\color{red}80.0}&\textbf{\color{red}72.1}&76.4&\textbf{\color{red}74.2}&\textbf{\color{red}72.4}&\textbf{\color{red}56.5}&41.4&\textbf{\color{blue}47.7}&71.9&\textbf{\color{blue}67.9}&87.6&\textbf{\color{red}76.5}\\
      \hline
  \end{tabular}}
    \caption{Comparing our ZeroMamba with CNN-based and ViT-based methods on CUB, SUN, and AWA2 benchmark datasets in the CZSL and GZSL settings. Our ZeroMamba significantly sets a new state-of-the-art (SOTA) for zero-shot learning. The best and second-best results are highlighted in \textbf{\color{red}Red} and \textbf{\color{blue}Blue}, respectively. Symbol ``--" denotes no results are reported.}
  \label{tab:sota}
\end{table*}

\section{Experiments}
\noindent{\textbf{Benchmark Datasets.}} To evaluate the effectiveness of our proposed framework, we conduct extensive experiments on three prominent ZSL datasets: Caltech-USCD Birds-200-2011 (\textbf{CUB}) \cite{Welinder2010CaltechUCSDB2}, SUN Attribute (\textbf{SUN}) \cite{patterson2012sun} and Animals with Attributes 2 (\textbf{AWA2}) \cite{xian2019f}. We use the Proposed Split (PS) \cite{xian2019f} division, as detailed in Tab. \ref{tab:dataset}. Additionally, we verify the generalization of ZeroMamba on large-scale ImageNet \cite{russakovsky2015imagenet} benchmark.
\begin{table}[htbp]
  \centering
  \vspace{-1mm}
      \begin{tabular}{c|c|c|c|c}
        \hline
        Datasets & Type &Images  & $N_{S}$ &  Classes ($s$ $|$ $u$)\\
        \hline
        \hline
        {\bf CUB}  &bird& 11788 & 312 & 200 (150 $|$ 50) \\
        {\bf SUN}  &scene& 14340 & 102& 717 (645 $|$ 72)  \\
        {\bf AWA2} &animal& 37322&85 &50 (40 $|$ 10)\\
        \hline
    \end{tabular}
      \caption{Details of the ZSL datasets. $N_{S}$ denotes the dimensions of the semantic vector. $s$ and $u$ are the number of seen and unseen classes.}
      \label{tab:dataset}
       \vspace{-5mm}
\end{table}\\

\noindent{\textbf{Evaluation Metrics.}} Following previous work \cite{xian2019f,naeem2024i2dformer+,chen2024progressive}, we measure the average Top-1 accuracy per unseen class for the CZSL setting, denoted as $Acc$. In the GZSL setting, we calculate the harmonic mean among seen and unseen classes: $H = (2~\times~S~\times~U)~/~(S~+~U)$. $S$ and $U$ indicate the Top-1 accuracy of seen and unseen classes.\\

\noindent{\textbf{Implementation Details.}}
We extend the first Vision Mamba-based model for ZSL, which differs from previous CNN-based and ViT-based models. Specifically, we leverage the VMamba-Small model pre-trained on ImageNet-1K to initialize our Vision Mamba Encoder. We implement our experiments in PyTorch and utilize the SGD optimizer (momentum $=$ 0.9, weight decay $=$ 0.001) with learning rate of $5\times{10^{-4}}$ on a single NVIDIA A100 GPU. All models are trained with a batch size of 16. In our method, we empirically set $\{\lambda_{sc}, \lambda_{col}\}$ to \{1.0,0.3\}, \{0.2,0.35\}, and \{0.0,0.98\} for CUB, SUN, and AWA2, respectively.

\subsection{Comparison with State-of-the-Art Methods}
\vspace{-1mm}
We compare the proposed ZeroMamba with SOTA CNN-based and ViT-based methods, with the results shown in Tab. \ref{tab:sota}. In the CZSL task, ZeroMamba achieves the best accuracy of 80.0\% and 72.4\% on CUB and SUN, respectively. For the coarse-grained dataset (\ie, AWA2), ZeroMamba remains competitive ($Acc$ = 71.9\%). For the challenging GZSL task, ZeroMamba surpasses the previous best methods (\eg, ZSLViT \cite{chen2024progressive} on CUB and I2DFormer+ \cite{naeem2024i2dformer+} on AWA2) that use ViT-Base backbone with more parameters. Additionally, compared to large-scale vision-language models (\eg, CLIP \cite{radford2021learning}), our method improves $H$ by 19.2\% on CUB. It is worth noting that ZeroMamba significantly improves the accuracy of unseen classes while maintaining superior performance on seen classes. Furthermore, the results in Fig. \ref{fig:motivate} strongly demonstrate that ZeroMamba attains more efficient computation than CNN-based and ViT-based methods, \ie, superior performance with fewer parameters. Overall, these results show the effectiveness and efficiency of our ZeroMamba in both the CZSL and GZSL settings. 
\begin{table*}[htbp]
\small
\centering
\resizebox{0.98\linewidth}{!}{
\begin{tabular}{l|c|c|c|c|c|c}
\hline
Method&SEKG\shortcite{wang2018zero}&CADA-VAE\shortcite{schonfeld2019generalized}&Transzero++ \shortcite{chen2022transzero++}&I2DFormer \shortcite{naeem2022i2dformer}&I2DFormer+ \shortcite{naeem2024i2dformer+}&\bf{ZeroMamba (Ours)}\\
\hline
$Acc$ (\%)&10.8&9.8&\textbf{\color{blue}23.9}&15.5&17.6&\textbf{\color{red}24.5}\\
\hline
\end{tabular}}
\caption{Comparison results on ImageNet. The best and second-best results are highlighted in \textbf{\color{red}Red} and \textbf{\color{blue}Blue}, respectively.} 
\label{table:imagenet}
\end{table*}

\begin{table*}[htbp]
\small
  \centering
  \resizebox{\linewidth}{!}{
  \begin{tabular}{l|c|c|ccc|c|ccc|c|ccc}
  \hline
      \multirow{3}{*}{\textbf{Model}}  &\multirow{3}{*}{\textbf{Params (M)}}&\multicolumn{4}{c|}{\textbf{CUB}}&\multicolumn{4}{c|}{\textbf{SUN}}&\multicolumn{4}{c}{\textbf{AWA2}}\\
      \cline{3-14}
      &&\multicolumn{1}{c|}{CZSL} & \multicolumn{3}{c|}{GZSL}&\multicolumn{1}{c|}{CZSL} & \multicolumn{3}{c|}{GZSL}&\multicolumn{1}{c|}{CZSL} & \multicolumn{3}{c}{GZSL}\\
      \cline{3-14} 
      &&\multicolumn{1}{c|}{\rm{Acc}} & \rm{U} & \rm{S} &\rm{H} & \multicolumn{1}{c|}{\rm{Acc}} & \rm{U} &\rm{S} & \rm{H} & \multicolumn{1}{c|}{\rm{Acc}} &\rm{U}  & \rm{S}  & \rm{H}\\
      \hline
      LocalVim-Tiny \shortcite{huang2024localmamba}&8.3&66.5&55.8&	60.4&	58.0&62.6&42.3&	33.3&	37.3&64.3&58.9&	76.6&	66.6\\
      Vim-Small \shortcite{zhu2024vision}&25.9&64.7&58.2&	52.1&	55.0&64.4&58.5&	21.2&	31.2&62.2&58.2&	79.1&	67.1\\
      LocalVim-Small \shortcite{huang2024localmamba}&27.9&68.1&56.6&	63.7&	59.9&65.6&46.2&	36.4&	40.7&66.7&53.2&	84.4&	65.2\\
      LocalVMamba-Small \shortcite{huang2024localmamba}&50.1&74.0&64.6&	64.3&	64.4&69.4&51.5&	37.4&	43.3&64.2&55.9&	84.4&	67.3\\
      \rowcolor{gray!20}
      \textbf{ZeroMamba-Tiny (Ours)}&31.4&78.4&71.6&	74.3&	72.9&69.9&52.0&	41.1&	45.9&67.7&61.2&	87.2&	71.9\\
      \rowcolor{gray!20} 
      \textbf{ZeroMamba-Small (Ours)}&51.3&\textbf{\underline{80.0}}&72.1&	76.4&	\textbf{\underline{74.2}}&\textbf{\underline{72.4}}&56.5&	41.4&	\textbf{\underline{47.7}}&\textbf{\underline{71.9}}&67.9&	87.6&	\textbf{\underline{76.5}}\\
      \hline
  \end{tabular}}
  \caption{Comparing our ZeroMamba with different Vision Mamba models on CUB, SUN, and AWA2 benchmark datasets in the CZSL and GZSL settings. The best result is highlighted in \textbf{boldface} and \underline{underline}.}
  \label{tab:different model}
\end{table*}

\begin{table*}[!t]
  \centering
  \resizebox{0.9\linewidth}{!}{
    \begin{tabular}{c|c|c|c|c|ccc|c|ccc}
      \hline
      \multirow{3}{*}{baseline} & \multirow{3}{*}{$\mathcal{L}_{sc}$} & \multirow{3}{*}{SLP}& \multirow{3}{*}{GRL}&\multicolumn{4}{c|}{\textbf{CUB}}&\multicolumn{4}{c}{\textbf{SUN}}\\
      \cline{5-12}
      &&&&\multicolumn{1}{c|}{CZSL} & \multicolumn{3}{c|}{GZSL}&\multicolumn{1}{c|}{CZSL} & \multicolumn{3}{c}{GZSL}\\
      \cline{5-12} 
      &&&&\multicolumn{1}{c|}{\rm{Acc}} & \rm{U} & \rm{S} &\rm{H} & \multicolumn{1}{c|}{\rm{Acc}} & \rm{U} &\rm{S} & \rm{H}\\
      \hline
      \checkmark&\checkmark&&&75.7&73.2&62.9&67.6&70.6&49.5&41.3&45.1\\
      \checkmark&\checkmark&\checkmark&&76.2&73.0&65.4&69.0&71.3&46.0&45.6&45.8\\
      \checkmark&\checkmark&&\checkmark&76.9&73.7&65.8&69.6&71.8&55.4&40.0&46.5\\
      \checkmark&&\checkmark&\checkmark&77.6&59.0&80.6&68.1&71.2&50.3&43.4&46.6\\
      \rowcolor{gray!20}
      \textbf{ZeroMamba (full)}&\checkmark&\checkmark&\checkmark&\textbf{\underline{80.0}}&72.1&	76.4&	\textbf{\underline{74.2}}&\textbf{\underline{72.4}}&56.5&	41.4&	\textbf{\underline{47.7}}\\
      \hline
  \end{tabular}}
  \caption{Ablation studies for different components of ZeroMamba on CUB and SUN datasets. We directly remove the classification head of VMamba and then insert an MLP with $\mathcal{L}_{sc}$ to map visual features to the semantic space as our baseline. SLP and GRL represent the Semantic-aware Local Projection and the Global Representation Learning modules, respectively. The best result is highlighted in \textbf{boldface} and \underline{underline}.}
  \label{tab:ablation study}
\end{table*}

\subsection{Experiments on Large-scale Dataset}
\vspace{-1mm}
To further evaluate the effectiveness and generalization capabilities of ZeroMamba, we conduct experiments on a challenging large-scale benchmark (\ie, ImageNet). Due to the difficulty of this task, most recent work has excluded it. In this work, we follow TransZero++ \cite{chen2022transzero++} to obtain semantic embeddings of class names using word2vec \cite{mikolov2013efficient} with 300-dimensions. We randomly split the training/test set into 800/200. We only use 1/10 of the data for every class for training. The experimental results are in Tab. \ref{table:imagenet} for the CZSL task. We observe that ZeroMamba outperforms all methods, setting a new SOTA on ImageNet. Also, ZeroMamba exhibits an impressive zero-shot accuracy of 24.5\% compared to I2DFormer+ ($Acc =$ 17.6\%) \cite{naeem2024i2dformer+} which directly learns a class embedding from the document text with global and fine-grained alignment. These results indicate the superior generalization ability of ZeroMamba.

\subsection{Ablation Study and Analysis}
In this section, we conduct ablation studies and analyses to comprehensively demonstrate the effectiveness of our proposed ZeroMamba, including the effectiveness of different Vision Mamba models and components. Additionally, we verify the impact of different visual backbones and input sizes. Due to page limitations, detailed results and discussions on these aspects are provided in Appendices \textcolor{red}{B} and \textcolor{red}{C}.\\

\noindent{\textbf{Effectiveness of Different Vision Mamba Models.}} To further assess the effectiveness of our proposed ZeroMamba, we conduct experiments on different Vision Mamba models. The comparison results are shown in Tab. \ref{tab:different model}. We map these models' visual features (\ie, LocalVim, Vim and LocalVMamba) to the semantic space using an MLP for classification. Evidently, our ZeroMamba (using VMamba \cite{liu2024vmamba} as the visual encoder) achieves the best results on all datasets with only slightly increased parameters. Moreover, our method incorporates two semantic feature learning modules and fuses semantic information, demonstrating the rationality and effectiveness of our design.\\

\noindent{\textbf{Effectiveness of Each Component.}} To verify the effectiveness of our proposed components, we report the improvements of exerting components on our baseline model in Tab. \ref{tab:ablation study}. The results indicate that merely altering the classification head to a semantic-mapping MLP does not yield satisfactory performance (\ie, baseline). When ZeroMamba lacks $\mathcal{L}_{sc}$ to exert semantic constraint, the $Acc/H$ significantly drops by 4.3\%/6.1\% and 1.8\%/1.1\% on CUB and SUN, respectively. This is because fine-grained datasets can't ensure semantic consistency, leading to severe overfitting of the classification results to seen classes. Moreover, both SLP and GRL bring notable performance gains, verifying that enhancing global features and utilizing local features benefit discriminative semantic learning. As a result, the model achieves $Acc/H$ improvements of 4.3\%/6.6\% and 1.8\%/2.6\% on CUB and SUN, respectively. In general, these results indicate the effectiveness of each component of ZeroMamba.

\begin{figure*}[!ht]
  \centering
   \includegraphics[width=0.98\linewidth]{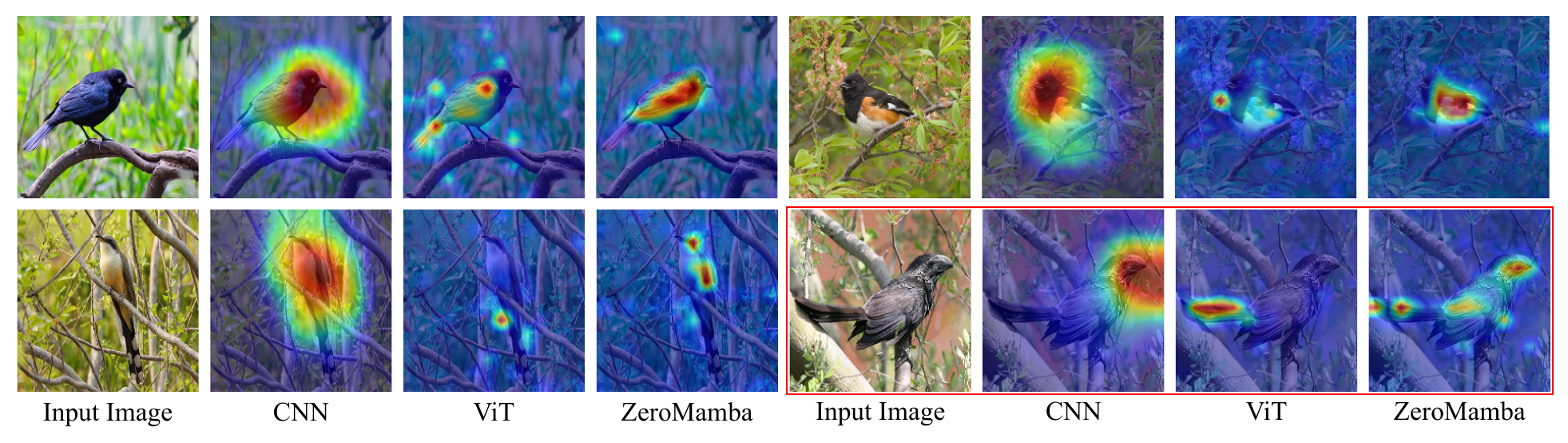}
   \caption{Visualization of the activation maps of different visual backbones, including CNN (\eg, ResNet-101 \cite{he2016deep}), ViT (\eg, ViT-Base \cite{dosovitskiy2020image}), and ZeroMamba. Our ZeroMamba can accurately capture the semantic-related information. We use CUB as an example, with the red box indicating challenging cases.} 
   \label{fig:atttention_cub}
\end{figure*}
\begin{figure}[!ht]
  \centering
   \includegraphics[width=1.0\linewidth]{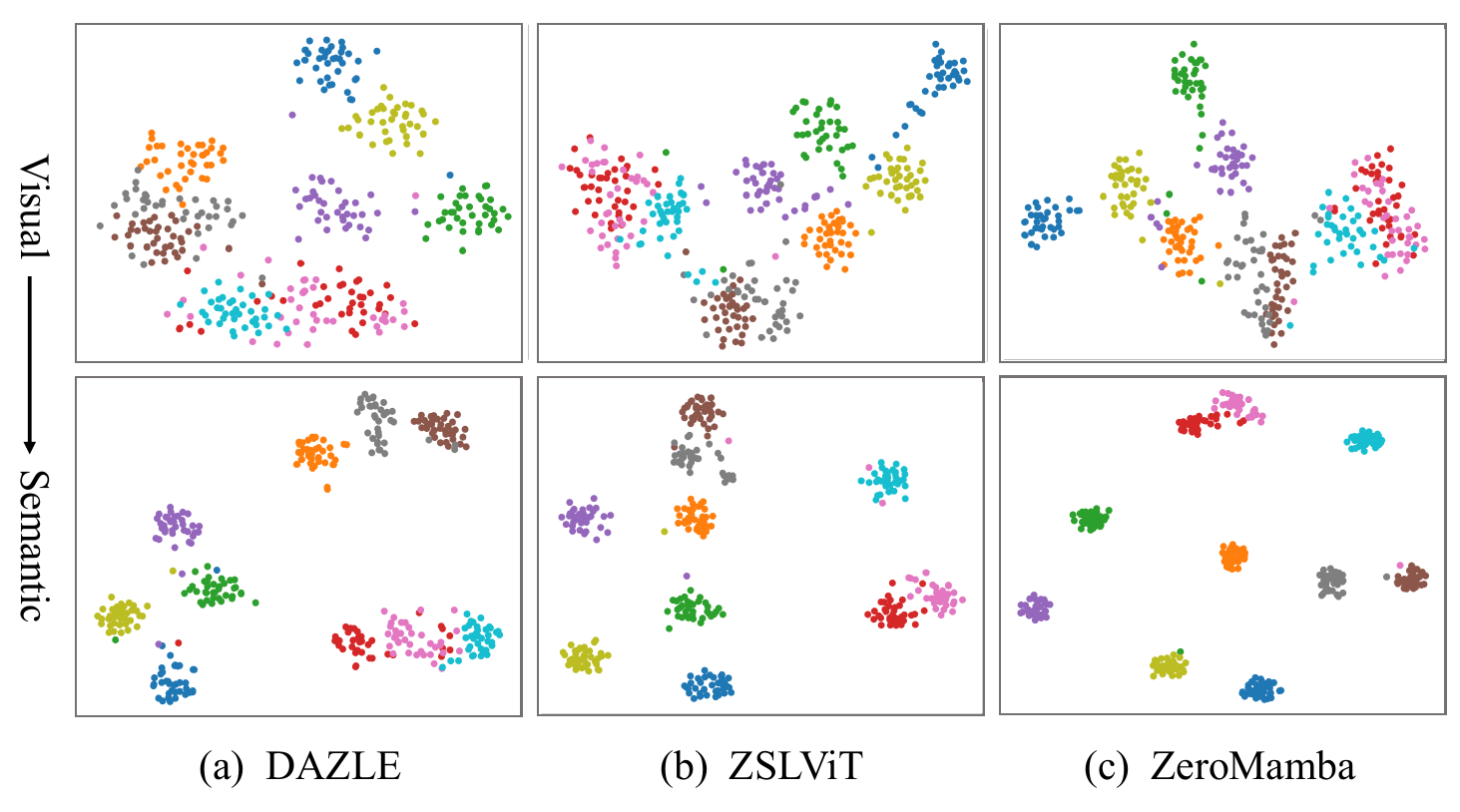}
   \caption{(Please Zoom in for details.) Visualizations with t-SNE embeddings of different methods produced by (a) DAZLE \cite{huynh2020fine}, (b) ZSLViT \cite{chen2024progressive}, and (c) ZeroMamba (ours) in both visual and semantic spaces. The 10 colors denote 10 different classes randomly selected from CUB. (Best viewed in shape and color.)} 
   \label{fig:tsne_cub_seen}
\end{figure}
\subsection{Qualitative Results}
\noindent{\textbf{Visualization of Activation Maps.}} We visualize the activation maps of our proposed ZeroMamba alongside pioneering visual backbones, as shown in Fig. \ref{fig:atttention_cub}. Obviously, CNN captures global representations, and ViT focuses on local object parts, while our ZeroMamba detects the semantic-related regions that are beneficial for classification. The part enclosed in a red rectangle displays a hard case where all three models perform suboptimally, yet ZeroMamba can still localize more semantic-related information. This shows the capability of ZeroMamba to handle complex visual-semantic relationships. More results are placed in the Appendix \textcolor{red}{D}.\\
\begin{figure}[!ht]
  \centering
   \includegraphics[width=1.0\linewidth]{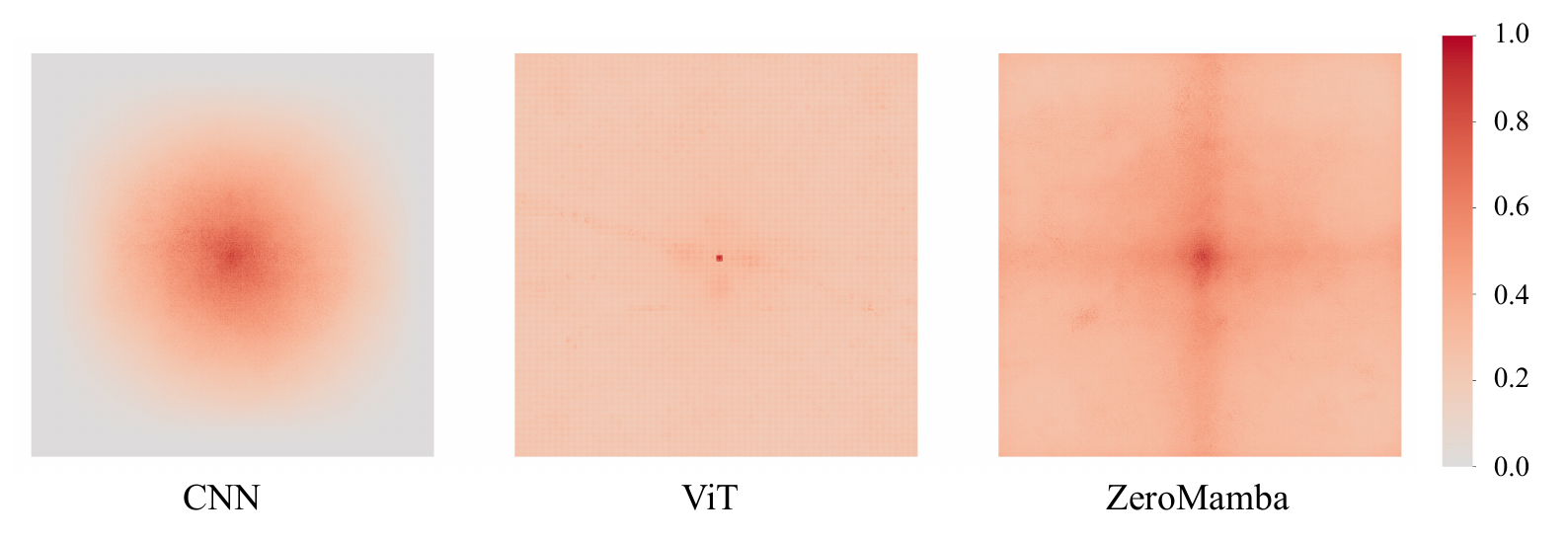}
   \caption{Comparison of effective receptive fields (ERF) \cite{luo2016understanding} between CNN (\eg, ResNet-101 \cite{he2016deep}), ViT (\eg, DeiT \cite{touvron2021training}) and ZeroMamba. Pixels with higher intensity (darker color) indicate a stronger response to the central pixel. It is evident that ZeroMamba and ViT exhibit a global receptive field, while CNN only has a local receptive field on CUB.} 
   \label{fig:cub_erf}
\end{figure}

\noindent{\textbf{Visualization of t-SNE Embeddings.}} We also present t-SNE embeddings \cite{van2008visualizing} to compare our ZeroMamba with classic CNN-based and ViT-based methods (\ie, DAZLE \cite{huynh2020fine} and ZSLViT \cite{chen2024progressive}). We randomly select 10 classes from CUB and visualize both visual and semantic features. As shown in Fig. \ref{fig:tsne_cub_seen}, it demonstrates that visual features are more scattered and mixed among classes, while semantic features are relatively cohesive and distinguishable. The comparison indicates that ZeroMamba optimizes visual and semantic representations, forming more distinct class boundaries in both visual and semantic spaces. These results strongly confirm that ZeroMamba is a desirable ZSL model. More t-SNE embeddings are shown in the Appendix \textcolor{red}{E}.\\

\noindent{\textbf{Visualization of Effective Receptive Fields.}} The Effective Receptive Field (ERF) \cite{luo2016understanding} plays a crucial role in visual tasks, as the output must respond to sufficiently large areas to capture semantic-related information from the entire image. In this work, we intuitively visualize the central pixel's ERF across different visual backbones. As shown in Fig. \ref{fig:cub_erf}, we can observe that CNN (\eg, ResNet-101 \cite{he2016deep}) exhibits local ERF, while ViT (\eg, DeiT \cite{touvron2021training}) and ZeroMamba demonstrate global ERFs. Furthermore, ZeroMamba shows higher intensity for the central pixel compared to ViT, indicating its superior capability in representing visual data. For more visualizations and discussions on SUN and AWA2, refer to Appendix \textcolor{red}{F}.

\begin{figure*}[ht]
    \begin{center}
    \subfigure[$\lambda_{sc}$ on CUB \hspace{-0.5 cm}]{
        \label{hyper:a}
    \includegraphics[width=.24\linewidth]{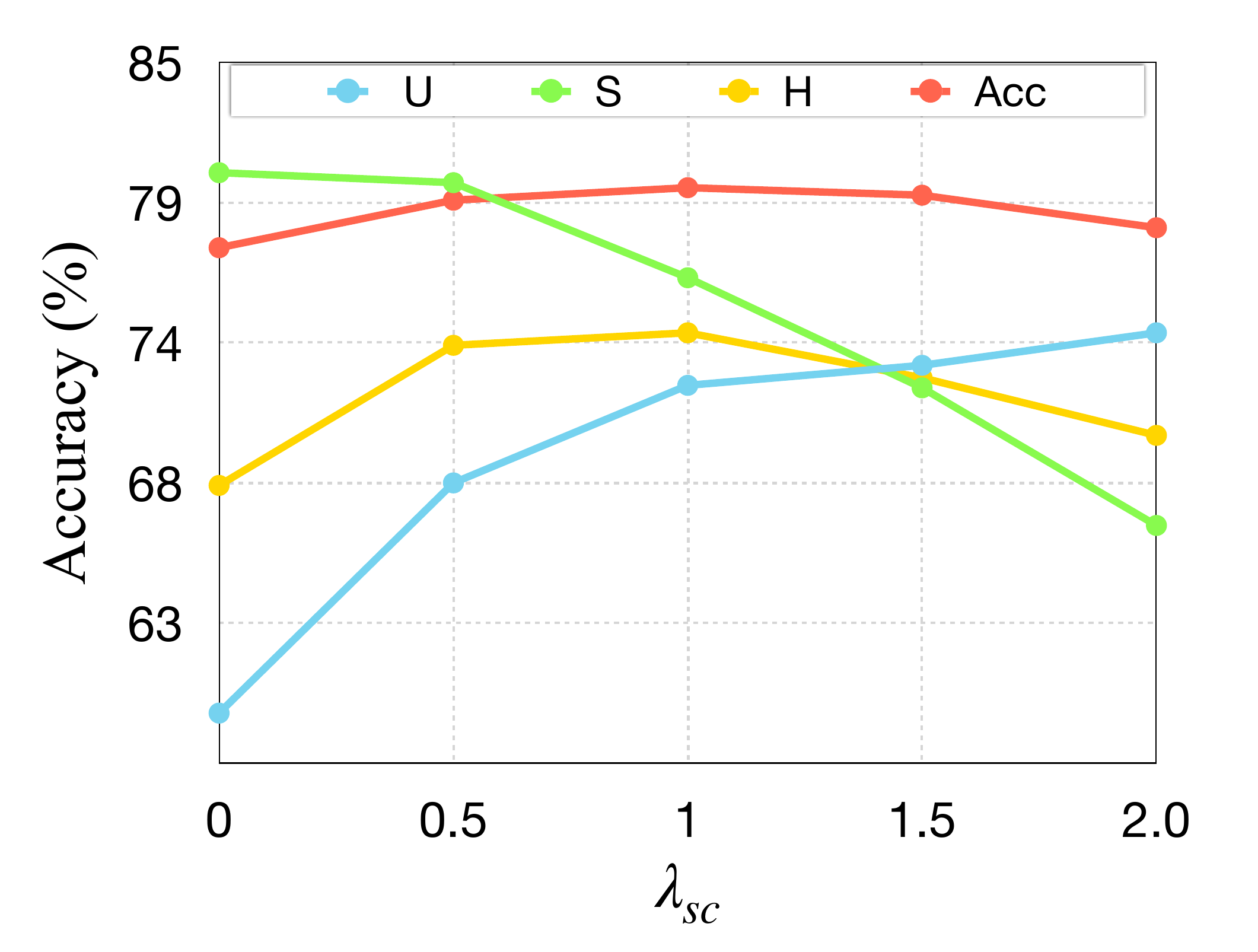}}
    \subfigure[$Batch~Size$ on CUB]{
        \label{hyper:b}
    \includegraphics[width=.24\linewidth]{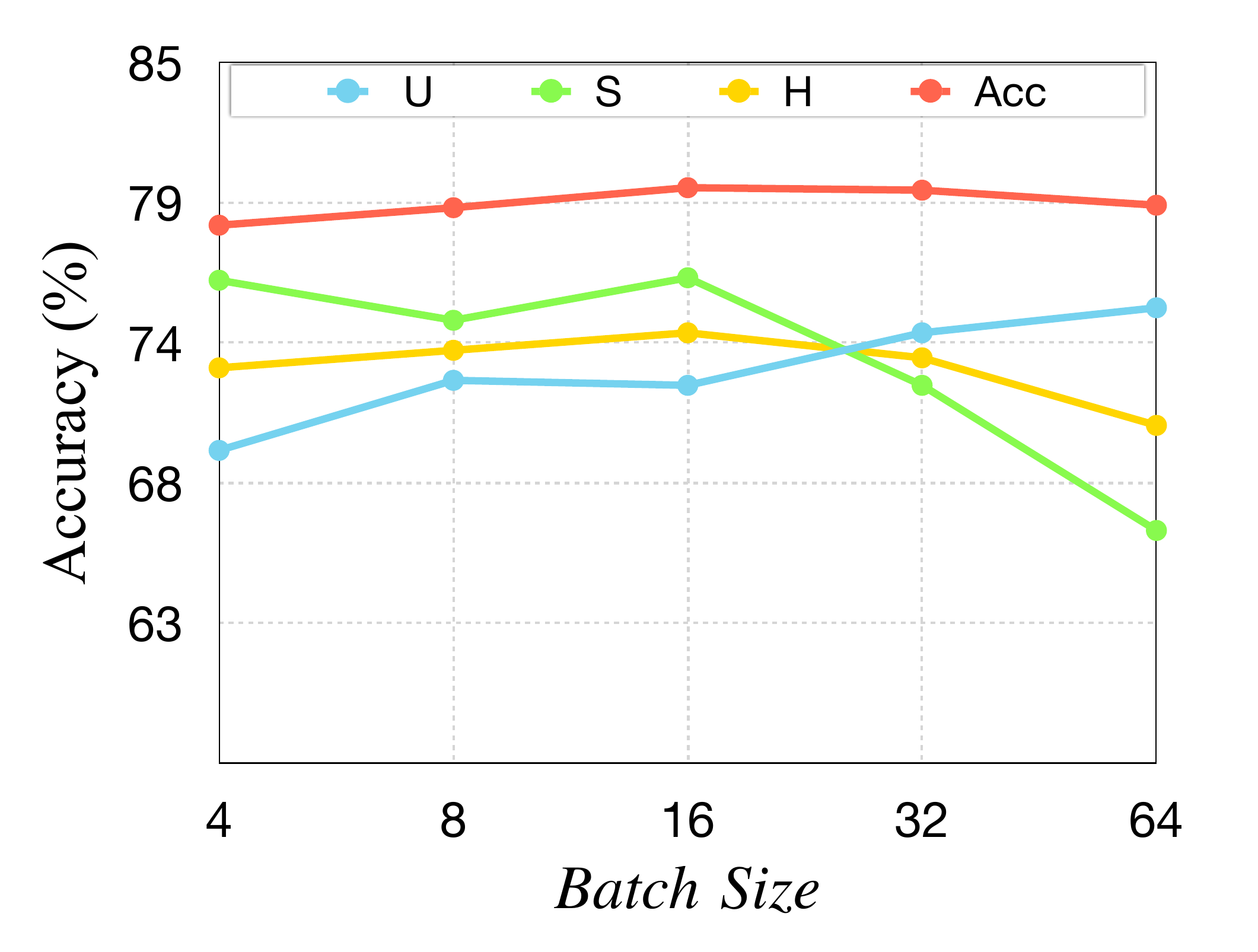}}
    \subfigure[$\lambda_{sc}$ on SUN \hspace{-0.5 cm}]{
        \label{hyper:c}
    \includegraphics[width=.24\linewidth]{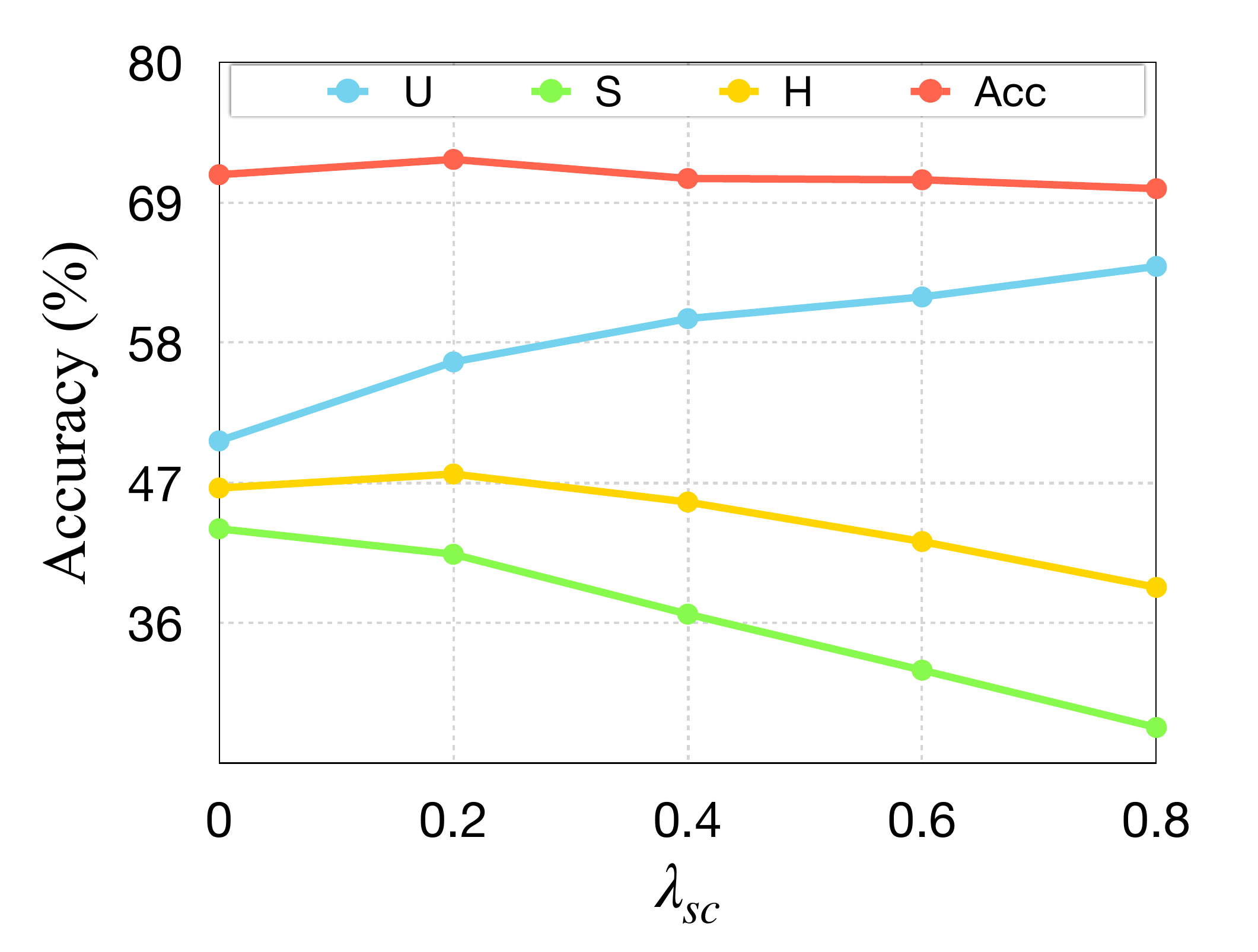}}
    \subfigure[$Batch~Size$ on SUN]{
        \label{hyper:d}
    \includegraphics[width=.24\linewidth]{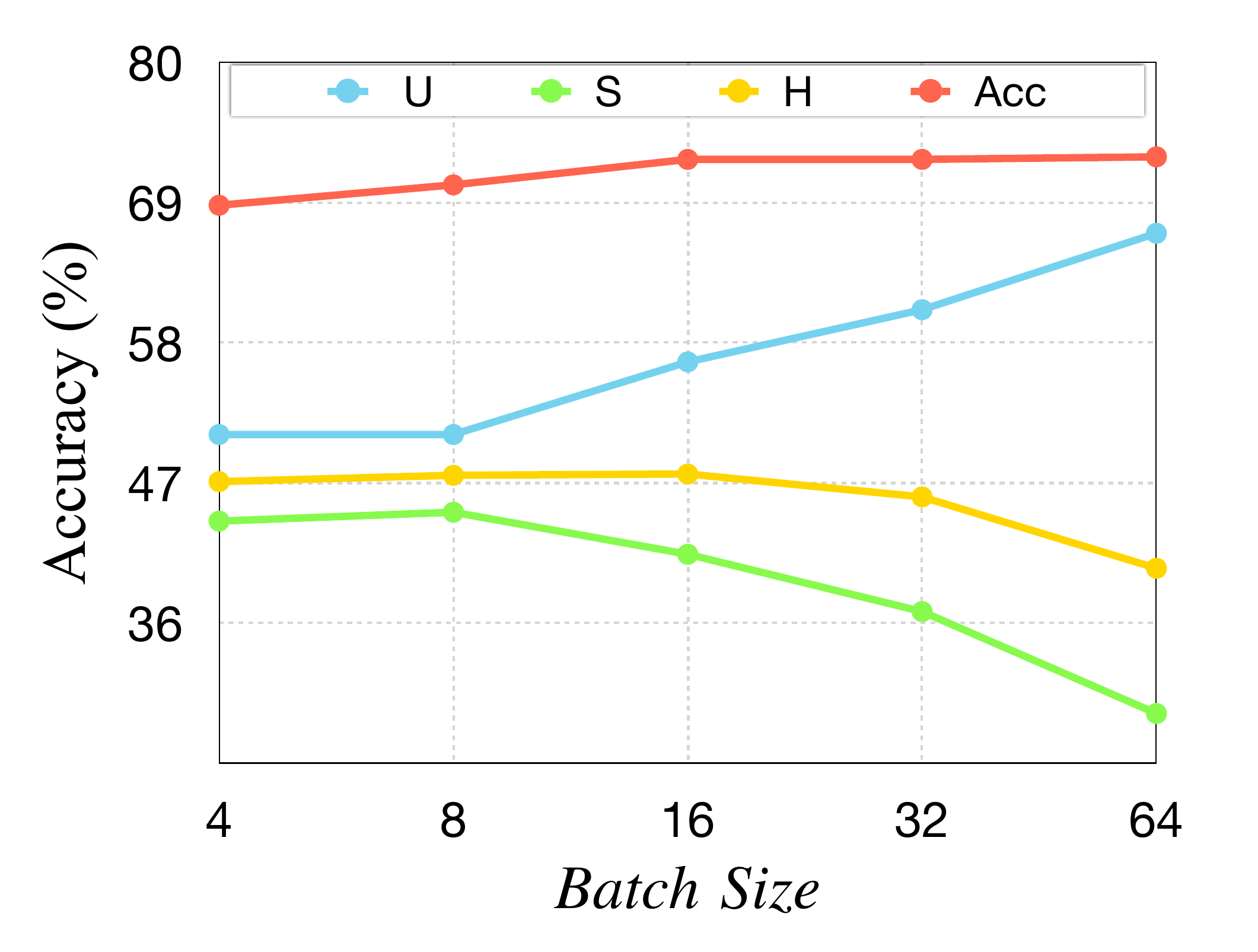}}
    \caption{Effect of (a) loss weight $\lambda_{sc}$ and (b) batch size on CUB and SUN.}
    \label{fig:hyper}
    \end{center}
  \end{figure*}
  
\begin{figure*}[!t]
    \begin{center}
    \subfigure[Calibration coefficient $\lambda_{col}$ on CUB \hspace{-0.5 cm}]{
        \label{fig:cub_com}
    \includegraphics[width=.46\linewidth]{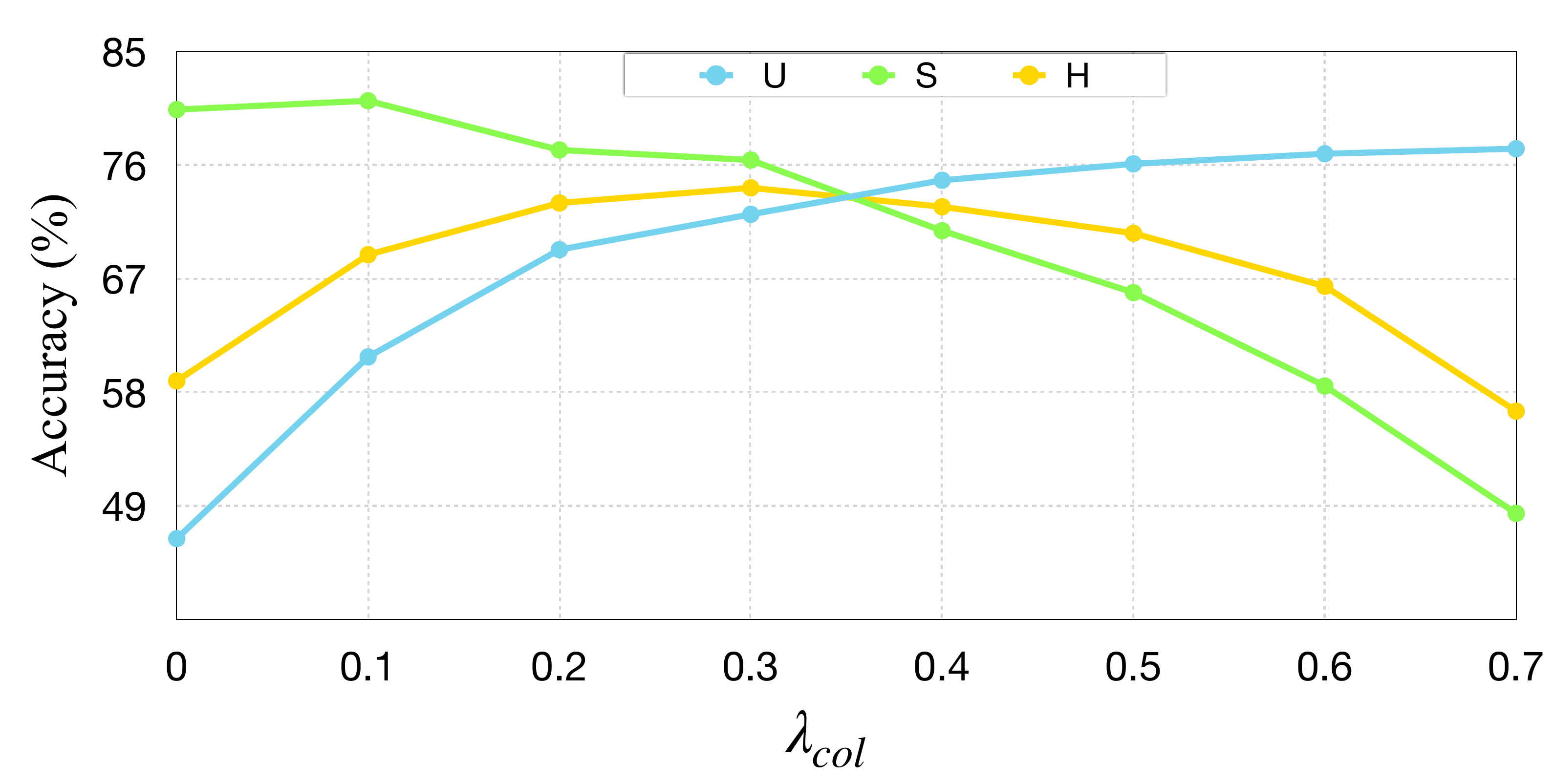}}
    \subfigure[Calibration coefficient $\lambda_{col}$ on SUN]{
    \includegraphics[width=.46\linewidth]{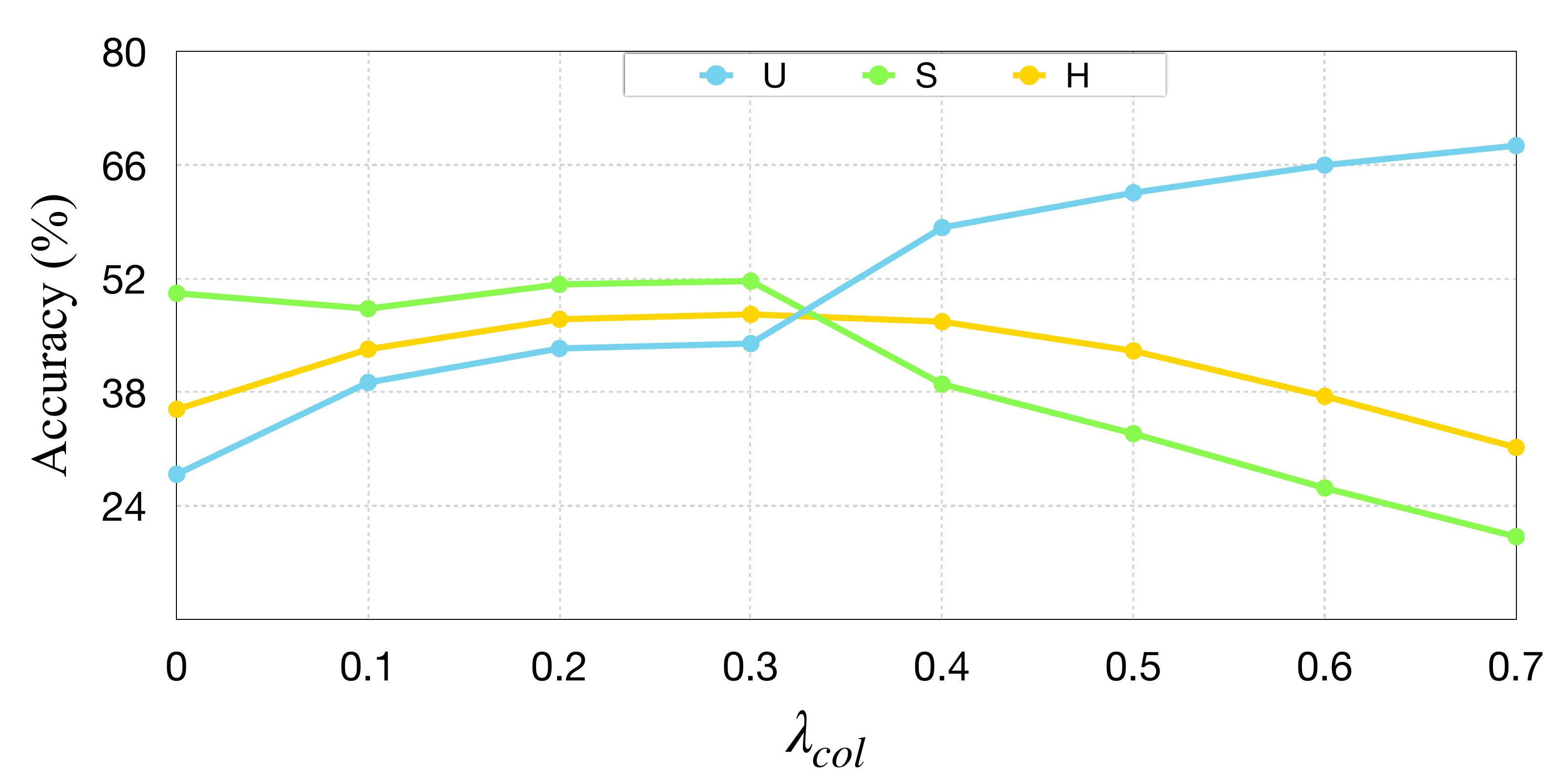}}
        \label{fig:sun_com}
    \caption{Effect of calibration coefficient $\lambda_{col}$ on CUB and SUN.}
    \label{fig:lambda}
        \end{center}
\end{figure*}

\subsection{Hyper-Parameter Analysis} 
We further analyze the impact of different hyper-parameters of our ZeroMamba on the CUB and SUN datasets. These hyper-parameters include the weight $\lambda_{sc}$ of the semantic constraint loss $\mathcal{L}_{sc}$, the $Batch\ Size$ for training, and the calibration coefficient $\lambda_{col}$. Fig. \ref{hyper:a} and Fig. \ref{hyper:c} show that an increase in $\mathcal{L}_{sc}$ leads to a higher $\textit{U}$, which indicates the efficacy of $\mathcal{L}_{sc}$ in reducing overfitting to seen classes. When $\mathcal{L}_{sc}$ is set to 1.0 and 0.2, $Acc/H$ obtains the maximum value on CUB and SUN. In Fig. \ref{hyper:b} and Fig. \ref{hyper:d}, our ZeroMamba indicates relative insensitivity to different batch sizes between different datasets. For all datasets, we set the batch size to 16. For $\lambda_{col}$, as shown in Fig. \ref{fig:lambda}, its innate effect is calibrating the bias against seen classes. A suitable $\lambda_{col}$  minimizes the gap between seen and unseen classes. Thus, to achieve the maximum $H$, we balance $S$ and $U$ with a relatively small value (\eg, 0.3 and 0.35 for CUB and SUN). Overall, ZeroMamba selects reasonable hyperparameters and is robust to them.

\section{Concluding Remarks}
In this paper, we have advocated Vision Mamba for ZSL, a parameter-efficient framework called ZeroMamba. The method is straightforward: incorporating semantic information into network learning and discovering local and global discriminative semantic-related representations to form an end-to-end ZSL framework. Our main contribution shows that despite its simplicity, ZeroMamba consistently outperforms existing CNN-based and ViT-based methods that have been developed specifically for ZSL and are more complex. Thus, this paper provides strong evidence that ZeroMamba can serve as an alternative baseline for ZSL. Furthermore, our investigation highlights the strengths of Vision Mamba in handling complex visual-semantic interactions, offering valuable insights for future research and improvements in the promising ZSL field. 

\bibliography{aaai25.bib}

\newpage
\section{Appendix}
This appendix is organized as: 
\begin{itemize}
\item Appendix {\color{red}A}: Details of the Vision Mamba Encoder.
\item Appendix {\color{red}B}: Results of Different Vision Backbone on CUB, SUN and AWA2.
\item Appendix {\color{red}C}: Results of Different Image Size on CUB, SUN and AWA2.
\item Appendix {\color{red}D}: Visualization of the Activation Maps on SUN and AWA2.
\item Appendix {\color{red}E}: Visualization of the t-SNE Embeddings on CUB, SUN and AWA2.
\item Appendix {\color{red}F}: Visualization of the Effective Receptive Fields on SUN and AWA2.
\end{itemize}

\begin{figure}[ht]
  \centering
   \includegraphics[width=1.0\linewidth]{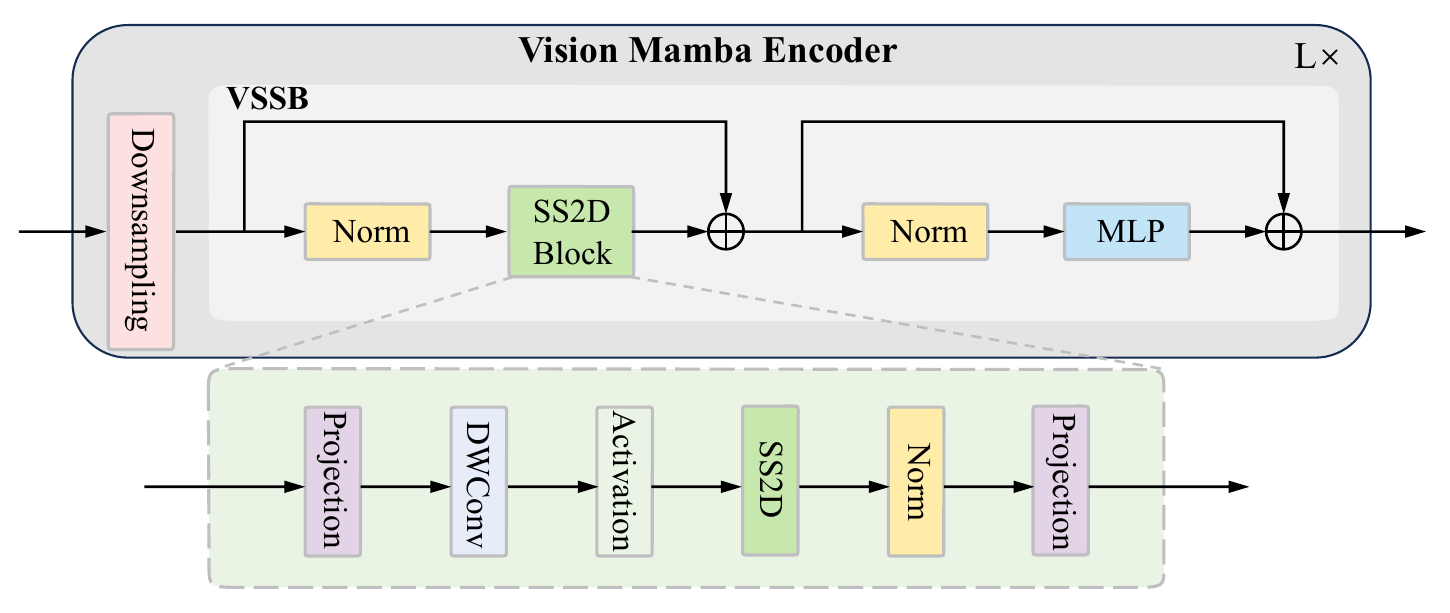}
   \caption{The structure of the Vision Mamba Encoder.}
   \label{fig:mamba_encoder}
\end{figure}

\noindent {\bf Appendix {\color{red}A}: Details of the Vision Mamba Encoder}\\
As shown in Fig. \ref{fig:mamba_encoder}, the Vision Mamba Encoder consists of alternating DownSampling blocks and visual state space blocks (VSSBs), where a single VSSB incorporates a 2D-Selective-Scan (SS2D) block and an MLP block with LayerNorm (Norm) employed before every block and residual connections after every block. In ZeroMamba, an input image $x_i$ is partitioned into patches $\mathcal{P}\in\mathbb{R}^{\frac{H}{4}\times{\frac{W}{4}}\times{C}}$, where $C$ is set to 96 in the VMamba-Small configuration. These patches are then processed by the vision Mamba encoder, yielding an output feature map $\mathcal{F}\in\mathbb{R}^{d_v\times{r}\times{r}}$. For more details, refer to VMamba \cite{liu2024vmamba}.\\

\noindent {\bf Appendix {\color{red}B}: Results of Different Vision Backbone on CUB, SUN and AWA2}\\
As shown in Tab. \ref{table:diff_bb}, we extend experiments to different vision backbones,  including DeiT \cite{touvron2021training}, ViT \cite{dosovitskiy2020image}, and VMamba \cite{liu2024vmamba}. 
For each visual backbone, we use an MLP to map visual features to the semantic space for classification. The results consistently demonstrate Vision Mamba's superior efficacy in ZSL tasks. Also, the experimental results indicate that ZeroMamba, compared to meticulously designed ViTs, can serve as an alternative model for the visual backbone, demonstrating strong visual representation capabilities.\\
\begin{table*}[htbp]
  \centering
  \resizebox{\linewidth}{!}{
  \begin{tabular}{l|c|ccc|c|ccc|c|ccc}
  \hline
      \multirow{3}{*}{\textbf{Backbone}}  &\multicolumn{4}{c|}{\textbf{CUB}}&\multicolumn{4}{c|}{\textbf{SUN}}&\multicolumn{4}{c}{\textbf{AWA2}}\\
      \cline{2-13}
      &\multicolumn{1}{c|}{CZSL} & \multicolumn{3}{c|}{GZSL}&\multicolumn{1}{c|}{CZSL} & \multicolumn{3}{c|}{GZSL}&\multicolumn{1}{c|}{CZSL} & \multicolumn{3}{c}{GZSL}\\
      \cline{2-13} 
      &\multicolumn{1}{c|}{\rm{Acc}} & \rm{U} & \rm{S} &\rm{H} & \multicolumn{1}{c|}{\rm{Acc}} & \rm{U} &\rm{S} & \rm{H} & \multicolumn{1}{c|}{\rm{Acc}} &\rm{U}  & \rm{S}  & \rm{H}\\
      \hline
      DeiT-Small \cite{touvron2021training}&74.1&65.9&	65.2&	65.5&63.8&48.7&	32.6&	39.1&63.7&56.5&	85.1&	67.9\\
      DeiT-Base \cite{touvron2021training}&74.3&64.0&	74.0&	68.7&66.9&49.2&	41.9&	45.3&68.8&63.8&	84.5&	72.7\\
      ViT-Base \cite{dosovitskiy2020image} &72.8&66.2&	66.3&	66.3&72.4&57.4&	40.1&	47.2&69.5&65.2&	86.8&	74.4\\
      VMamba-Small \cite{liu2024vmamba}&75.7&73.2&62.9&  67.6&70.6&49.5&41.3&45.1&66.4&58.3	&88.4&	70.2\\ 
      \rowcolor{gray!20}
      \textbf{ZeroMamba-Small (Ours)}&\textbf{\underline{80.0}}&72.1&	76.4&	\textbf{\underline{74.2}}&\textbf{\underline{72.4}}&56.5&	41.4&	\textbf{\underline{47.7}}&\textbf{\underline{71.9}}&67.9&	87.6&	\textbf{\underline{76.5}}\\
      \hline
  \end{tabular}}
    \caption{Results of different vision backbones on CUB, SUN, and AWA2 benchmark datasets in the CZSL and GZSL settings. The best result is highlighted in \textbf{boldface} and \underline{underline}.}
  \label{table:diff_bb}
  \end{table*}

\noindent {\bf Appendix {\color{red}C}: Results of Different Image Size on CUB, SUN and AWA2}\\
In Tab. \ref{tab:different input size}, we conduct experiments on three ZSL benchmark datasets with various image sizes (\ie, 224~$\times$~224, 384~$\times$~384 and 448~$\times$~448). As the results show, a larger image size somewhat boosts performance. Therefore, the input image size is 448~$\times$~448 in this work. Of course, our performance is also competitive at other resolutions.\\
\begin{table*}[htbp]
  \centering
  \resizebox{\linewidth}{!}{
  \begin{tabular}{l|c|c|ccc|c|ccc|c|ccc}
  \hline
      \multirow{3}{*}{\textbf{Method}} & \multirow{3}{*}{\textbf{Image Size}}&\multicolumn{4}{c|}{\textbf{CUB}}&\multicolumn{4}{c|}{\textbf{SUN}}&\multicolumn{4}{c}{\textbf{AWA2}}\\
      \cline{3-14}
      &&\multicolumn{1}{c|}{CZSL} & \multicolumn{3}{c|}{GZSL}&\multicolumn{1}{c|}{CZSL} & \multicolumn{3}{c|}{GZSL}&\multicolumn{1}{c|}{CZSL} & \multicolumn{3}{c}{GZSL}\\
      \cline{3-14} 
      &&\multicolumn{1}{c|}{\rm{Acc}} & \rm{U} & \rm{S} &\rm{H} & \multicolumn{1}{c|}{\rm{Acc}} & \rm{U} &\rm{S} & \rm{H} & \multicolumn{1}{c|}{\rm{Acc}} &\rm{U}  & \rm{S}  & \rm{H}\\
      \hline
      ZeroMamba&224~$\times$~224&75.5&68.0&	70.6&	69.3&69.7&52.4&	38.5&	44.3&69.0&65.5&	82.1&72.9\\
      ZeroMamba&384~$\times$~384&79.2&71.5&	75.6&	73.5&72.0&56.3&	40.5&	47.2&\textbf{\underline{72.1}}&67.6&	87.6&	76.3\\
      \rowcolor{gray!20}
      ZeroMamba&448~$\times$~448 &\textbf{\underline{80.0}}&72.1&	76.4&	\textbf{\underline{74.2}}&\textbf{\underline{72.4}}&56.5&	41.4&	\textbf{\underline{47.7}}&71.9&67.9&	87.6&	\textbf{\underline{76.5}}\\
      \hline
  \end{tabular}}
  \caption{Results of different image sizes on CUB, SUN, and AWA2 benchmark datasets in the CZSL and GZSL settings. The best result is highlighted in \textbf{boldface} and \underline{underline}.}
  \label{tab:different input size}
\end{table*}

\begin{figure*}[htbp]
    \begin{center}
    \subfigure[SUN ]{
        \label{hyper:att_sun}
    \includegraphics[width=1.0\linewidth]{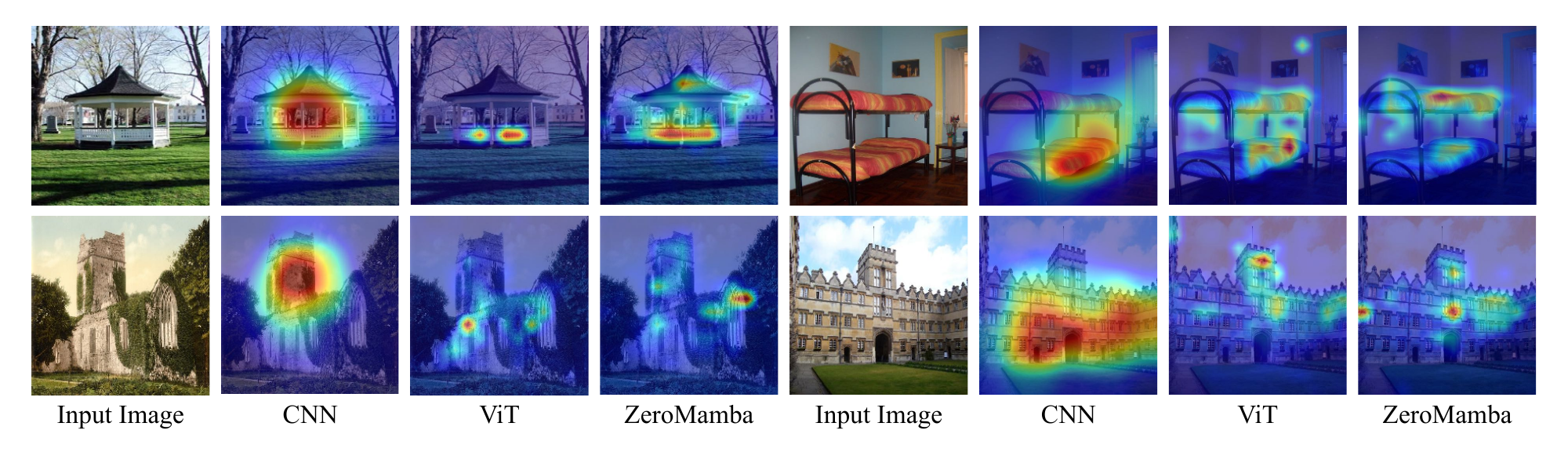}}
    \subfigure[AWA2]{
        \label{hyper:att_awa}
    \includegraphics[width=1.0\linewidth]{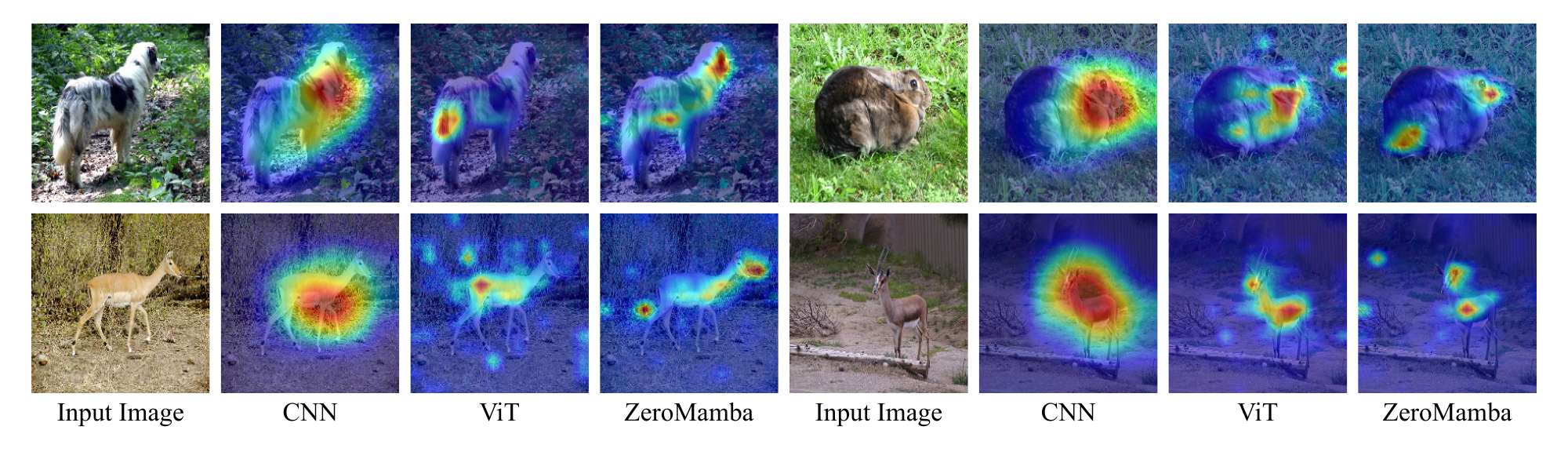}}
    \caption{Visualization of the activation maps of different visual backbones on SUN and AWA2, including CNNs (\eg, ResNet-101), ViTs (\eg, ViT-Base), and ZeroMamba. Our ZeroMamba can accurately capture the semantic-related information.} 
    \label{fig:cam_sun_awa}
    \end{center}
  \end{figure*}

\noindent {\bf Appendix {\color{red}D}: Visualization of the Activation Maps on SUN and AWA2}\\
Fig. \ref{fig:cam_sun_awa} visualizes the activation maps of our proposed ZeroMamba and previous visual backbones on the SUN and AWA2 datasets. For each dataset, we randomly select 4 samples. These visualizations demonstrate the significant advantage of ZeroMamba over CNNs and ViTs in capturing local semantic-related representations.\\

\begin{figure*}[!t]
    \begin{center}
    \subfigure[SUN\hspace{1.8cm}]{
        \label{hyper:tsne_sun}
    \includegraphics[width=0.95\linewidth]{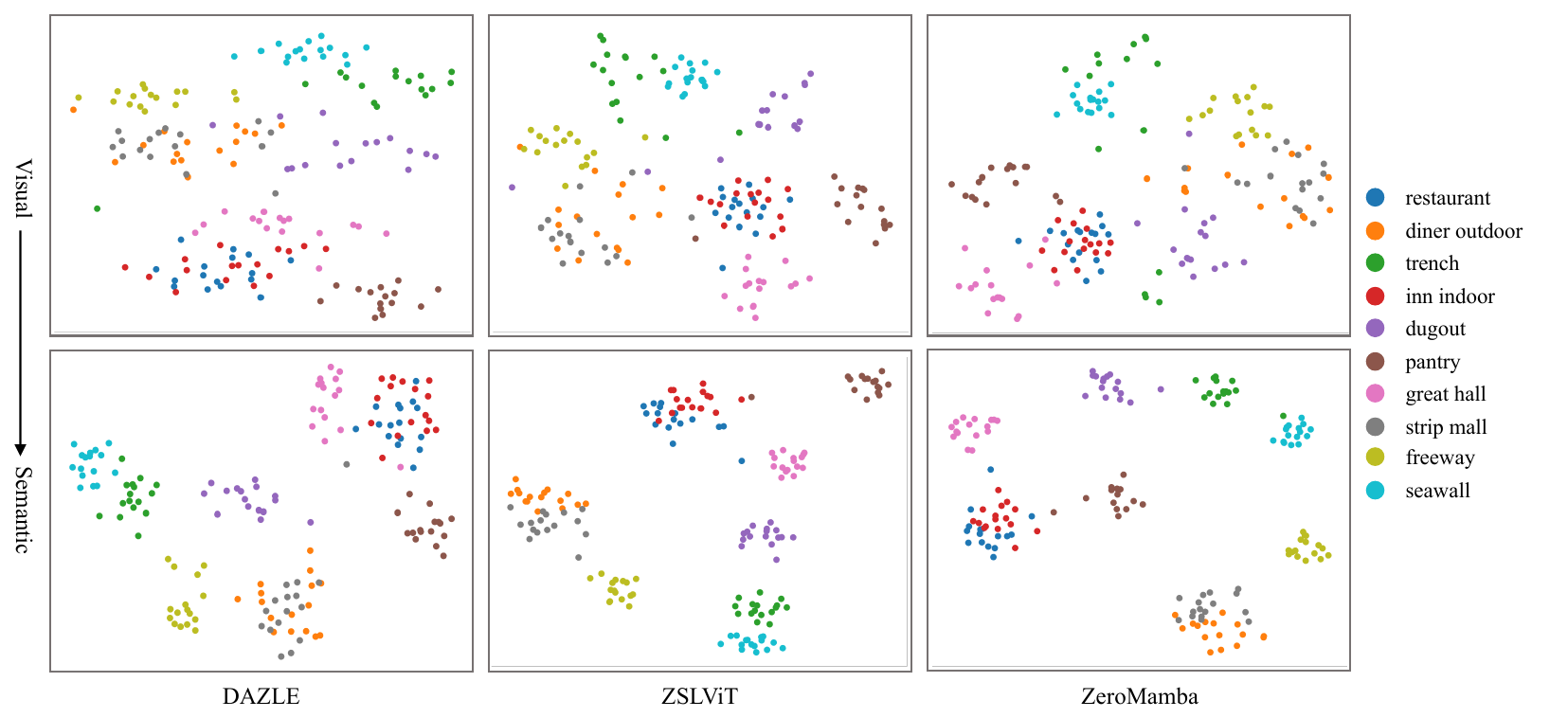}}
    \subfigure[AWA2\hspace{1.8cm}]{
        \label{hyper:tsne_awa}
    \includegraphics[width=0.95\linewidth]{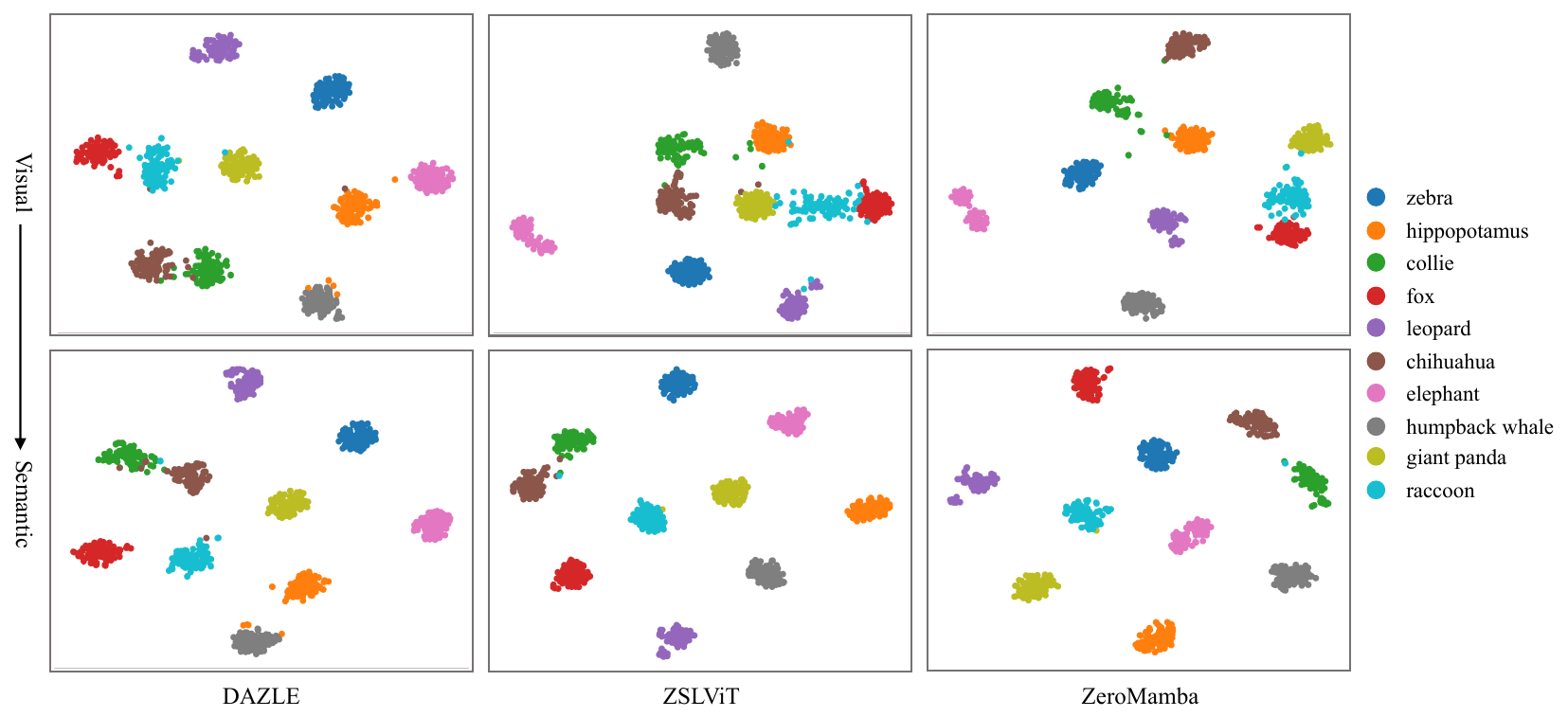}}
    \caption{Visualizations with t-SNE embeddings of different methods produced by DAZLE \cite{huynh2020fine}, ZSLViT \cite{chen2024progressive}, and ZeroMamba (ours) in visual and semantic spaces. The 10 colors denote 10 different classes randomly selected from SUN and AWA2.} 
    \label{fig:tsne_sun_awa}
    \end{center}
  \end{figure*}

\noindent {\bf Appendix {\color{red}E}: Visualization of the t-SNE Embeddings on SUN and AWA2}\\
Fig. \ref{fig:tsne_sun_awa} shows the t-SNE embedding visualizations of both visual and semantic features learned by DAZLE \cite{huynh2020fine}, ZSLViT \cite{chen2024progressive}, and ZeroMamba (ours) on SUN and AWA2. Analogously, the visual features tend to be more scattered, whereas the semantic features exhibit a more cohesive distribution. As a result, our ZeroMamba optimizes both visual and semantic representations, obtaining discriminative features that are better suited for ZSL. \\ 

\begin{figure*}[ht]
  \centering
   \includegraphics[width=0.78\linewidth]{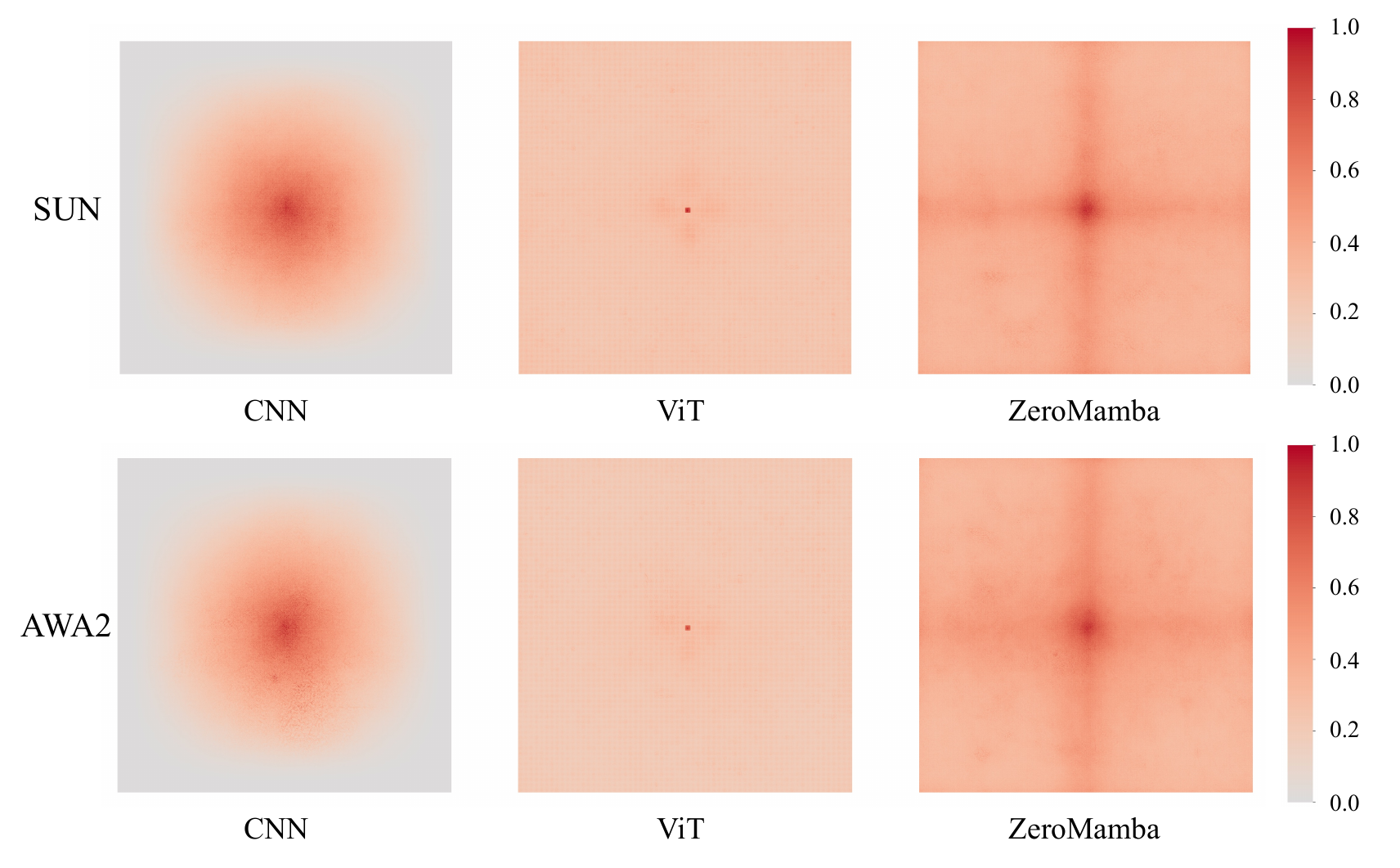}
   \caption{Comparison of effective receptive fields (ERF) between CNN (e.g., ResNet-101), ViT and ZeroMamba on SUN and AWA2. It is easy to observe that ZeroMamba and ViT exhibit a global receptive field, while CNN only has a local receptive field.}
   \label{fig:sun_awa2_erf}
\end{figure*}

\noindent{\bf Appendix {\color{red}F}: Visualization of the Effective Receptive Fields on SUN and AWA2}\\
The Effective Receptive Field (ERF) refers to the region containing any input pixel with a non-negligible impact on that unit. Fig. \ref{fig:sun_awa2_erf} visualizes the ERFs of different backbones in the SUN and AWA2 datasets. All models are visualized in a 1024 $\times$ 1024 image space after training. We observe that our ZeroMamba shows extensive coverage in its ERFs (\ie, fully filling the image) and a stronger response, demonstrating its efficacy in long-range modeling. Moreover, the linear time complexity of ZeroMamba makes it more computationally efficient than CNN and ViT.\\ 

\end{document}